\RequirePackage{xcolor}
\documentclass{article}

\let\EgoSteerAddContentsLine\addcontentsline

\usepackage{iclr2027_conference,times}
\let\addcontentsline\EgoSteerAddContentsLine

\usepackage{titletoc}

\titlecontents{section}
[1.6em]{\addvspace{0.7em}\bfseries}{\contentslabel{1.6em}}{}{\hfill\contentspage}
\titlecontents{subsection}
[4.0em]{}{\contentslabel{2.4em}}{}{\titlerule*[0.6pc]{.}\contentspage}
\titlecontents{subsubsection}
[7.2em]{}{\contentslabel{3.2em}}{}{\titlerule*[0.6pc]{.}\contentspage}

\usepackage{array}
\usepackage{tabularx}
\usepackage{hyperref}       
\usepackage{url}            
\usepackage{enumitem}
\usepackage{graphicx}
\usepackage{amsmath}
\usepackage{amssymb}
\usepackage{cleveref}
\usepackage[font=small,labelfont=bf]{caption}
\usepackage{wrapfig}
\usepackage{booktabs}
\usepackage{booktabs}
\usepackage{makecell}
\usepackage{relsize}
\usepackage{caption}
\usepackage{subcaption}
\usepackage{listings}
\usepackage{xcolor}
\usepackage{colortbl}
\usepackage{multirow}
\usepackage{listings}
\usepackage[most]{tcolorbox}
\usepackage{tikz}
\usepackage{pifont}
\usepackage{circledsteps}

\Crefname{appendix}{Appendix}{Appendices}

\definecolor{myblue}{HTML}{3B75C4}

\newcommand{\cmark}{\ding{51}}        
\definecolor{promptbg}{RGB}{248,248,248}
\definecolor{promptframe}{RGB}{215,215,215}
\definecolor{promptbar}{RGB}{80,120,160}
\definecolor{prompttitle}{RGB}{35,35,35}
\definecolor{promptrule}{RGB}{225,225,225}
\definecolor{zebra}{gray}{0.92}       

\lstdefinestyle{promptstyle}{
    basicstyle=\ttfamily\scriptsize,
    breaklines=true,
    breakatwhitespace=false,
    columns=fullflexible,
    keepspaces=true,
    showstringspaces=false,
    tabsize=2,
    frame=none,
    numbers=none,
    xleftmargin=0pt,
    xrightmargin=0pt,
    aboveskip=0pt,
    belowskip=0pt
}

\newtcblisting{breakableprompt}[2][]{
    listing only,
    listing options={style=promptstyle},
    breakable,
    enhanced jigsaw,
    colback=promptbg,
    colframe=promptframe,
    boxrule=0.35pt,
    sharp corners,
    borderline west={2pt}{0pt}{promptbar},
    left=8pt,
    right=7pt,
    top=7pt,
    bottom=7pt,
    before upper={
        {\sffamily\bfseries\small\color{prompttitle} #2}
        \par\vspace{4pt}
        {\color{promptrule}\hrule height 0.35pt}
        \vspace{6pt}
    },
    #1
}

\title{EgoSteer: An Open-Source Full-Stack System Towards Steerable Dexterous Manipulation from Egocentric Videos}

\author{%
Yifan Zhong\textsuperscript{1,2}\thanks{Equal contribution.\quad\textsuperscript{\textdagger}Corresponding authors.},
Zhang Chen\textsuperscript{1,2}\footnotemark[1],
Tianrui Guan\textsuperscript{1,2}\footnotemark[1],
Fanlian Zeng\textsuperscript{2,3}\footnotemark[1],
Ka Nam Lui\textsuperscript{1,2},\\
\bfseries Yuyao Ye\textsuperscript{1,2},
Tingrui Zhang\textsuperscript{1,2},
Jiayi Li\textsuperscript{1,2},
Tianjia He\textsuperscript{2},
Wenjie Lou\textsuperscript{1,2},
Ruilin Yan\textsuperscript{1,2},\\
\bfseries Xinhao Ji\textsuperscript{1,2},
Guangyu Zhao\textsuperscript{1,2},
Jiayuan Zhang\textsuperscript{1,2},
Wenxi Xu\textsuperscript{2,4},
Chengdong Ma\textsuperscript{1},\\
\bfseries Yuanpei Chen\textsuperscript{1,2}\footnotemark[2],
Yaodong Yang\textsuperscript{1,2}\footnotemark[2]\\
\normalfont\textsuperscript{1}Institute for AI, PKU, \textsuperscript{2}PKU-PsiBot Joint Lab,
\normalfont\textsuperscript{3}UPenn,
\normalfont\textsuperscript{4}SUSTech
}

\iclrfinalcopy

\begin{document}
\maketitle
\lhead{}


\vspace{-20pt}
\begin{abstract}
\vspace{-12pt}
The enduring vision of general-purpose robots serving humanity hinges fundamentally on policy \textit{steerability}. However, prevailing paradigms of learning from expert demonstrations demand massive real-world data even on simplified grippers, rendering them prohibitively expensive for high-dimensional, data-scarce dexterous hands. To overcome this bottleneck, we present a full-stack system that scales dexterous VLA pre-training from \textit{egocentric human videos} and enables data-efficient real-robot post-training. It integrates \textbf{EgoSmith}, a data pipeline that curates in-the-wild egocentric videos into $9,606$ hours of pre-training data with $8.3\times$ higher throughput and better accuracy than prior SOTA; a unified \textbf{Robot Stack} for teleoperation and human-in-the-loop correction tailored for dexterous hands; and \textbf{EgoSteer}, a world-model-enhanced VLA operating on a morphology-aligned action space. Human data pre-training equips EgoSteer with language-guided manipulation priors, which are grounded through robot post-training and further refined via DAgger. Empirically, EgoSteer robustly executes free-form instructions across 45 diverse tasks, demonstrating adherence to user intent amid multiple candidate tasks and generalization. The pre-trained model also few-shot adapts to five complex long-horizon tasks, including box folding, on two embodiments with 79\% average progress. All system code, datasets, model checkpoints, and an evaluation gallery are publicly available at \url{egosteer.github.io}.
\end{abstract}

\vspace{-15pt}

\begin{figure}[h]
  \centering
  \includegraphics[
  width=1.0\linewidth,
  trim={0bp 0bp 0bp 0bp},
  clip
]{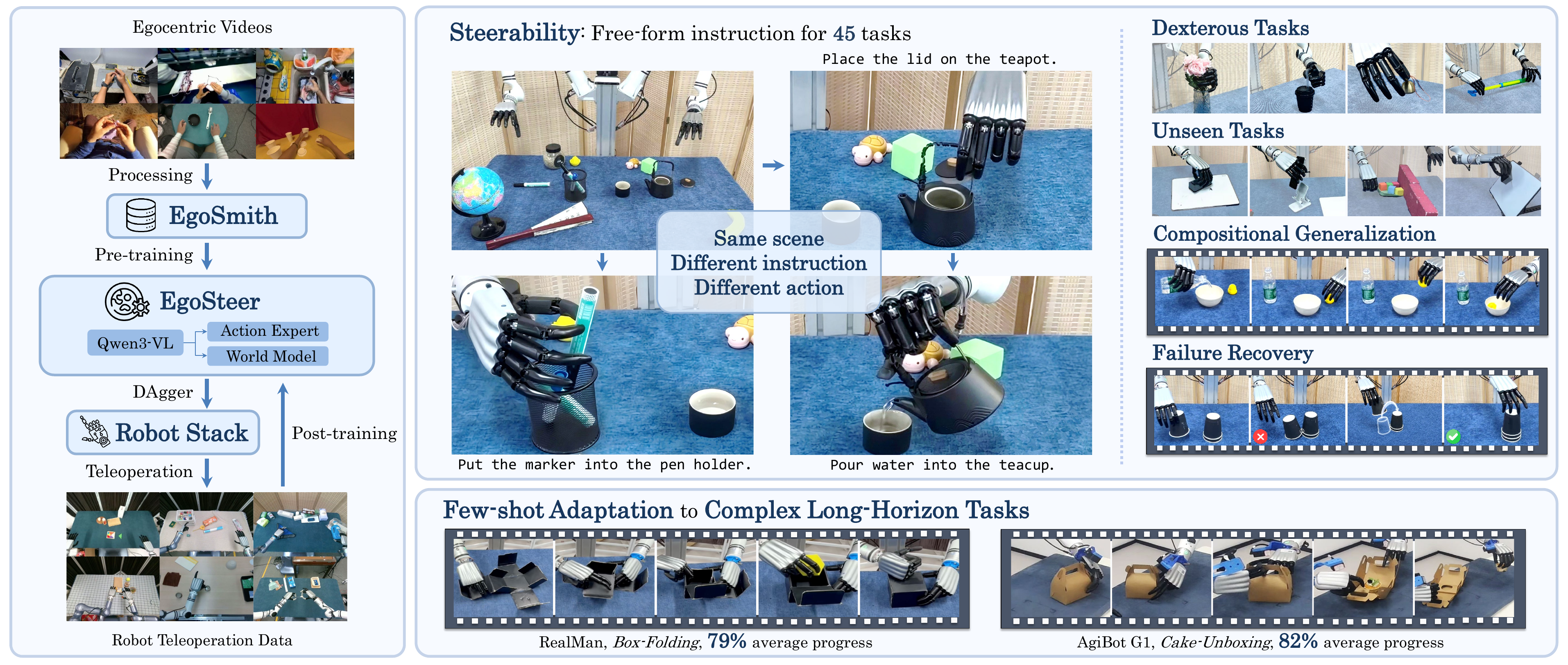}
\caption{Our \textbf{full-stack system} integrates EgoSmith, Robot Stack, and EgoSteer to learn from large-scale egocentric human videos and facilitate data-efficient real-robot post-training, enabling steerable dexterous manipulation across $45$ tasks as well as few-shot adaptation to complex, long-horizon tasks.}
\label{fig:teaser}
  \vspace{-10pt}
\end{figure}

\vspace{-8pt}
\section{Introduction}
\label{sec:introduction}
\vspace{-8pt}
A central goal of embodied intelligence is to enable robots to perform diverse manipulation tasks from open-ended human intent. Despite rapid progress in embodied foundation models~\citep{black2026pi0visionlanguageactionflowmodel, liu2025rdt, luo2025beingh0visionlanguageactionpretraininglargescale, lingbot-vla, brohan2023rt1roboticstransformerrealworld, zitkovich2023rt}, most systems still require task-specific fine-tuning, while the few that follow free-form language are largely limited to grippers~\citep{pi07, ye2026worldactionmodelszeroshot}. Dexterous hands provide a more expressive embodiment that offers greater actuation capacity and fine-grained interaction potential. Yet on such challenging platforms, steerable manipulation remains largely unrealized.

The key bottleneck lies in data and system scalability. While language-guided manipulation typically demands thousands of hours of data even on simplified grippers~\citep{pi07, agarwal2026cosmos, liu2026rdt2, lingbot-vla}, collecting sufficient expert demonstrations directly on high-dimensional, data-scarce dexterous hands is prohibitively expensive. 
Egocentric human videos~\citep{hoque2025egodex, punamiya2026egoverse} offer a scalable alternative, as human hand manipulation contains rich interaction knowledge and is abundant in the wild. However, raw egocentric videos are noisy and lack reliable language and action annotations. Without systematic curation, these unstructured videos yield unstable supervision that can degrade downstream robot policies. Even with curated human data, the system must employ high-capacity models trained with effective objectives, and ground the learned priors to the target robot. Failing to address any of these interdependent components prevents the realization of general language-following manipulation.

In this paper, we close this gap by proposing a full-stack system for steerable dexterous manipulation that scales VLA pre-training from egocentric human videos and enables data-efficient real-robot post-training. Our system begins with \textbf{EgoSmith}, a data pipeline that transforms in-the-wild video into clean, fully-annotated pre-training data. Through a tailored four-stage design, EgoSmith aligns RGB-D frames, world-space hand trajectories, camera parameters, and textual instructions, curating a $9.6$K-hour pre-training corpus across $12$ egocentric datasets. In particular, for 4D motion estimation, the pipeline's primary accuracy and computational bottleneck, we improve upon the open-source SOTA HaWoR~\citep{zhang2025hawor} to achieve higher precision and a $8.3\times$ throughput speedup. To ground these human-hand interaction priors onto physical embodiments, we design a unified \textbf{Robot Stack} for teleoperation, inference, and human-in-the-loop correction. With this stack, we collect $187$ hours of high-quality teleoperation data across $193$ semantically-diverse tasks for post-training. During deployment, it enables seamless expert intervention for efficient DAgger~\citep{ross2011reduction} refinement. Finally, we introduce \textbf{EgoSteer}, a world-model-enhanced VLA operating on a morphology-aligned action space. By incorporating a world-model expert that predicts action-induced future states in the DINOv3~\citep{simeoni2025dinov3} latent space, EgoSteer enhances the VLM backbone's action imagination and modality alignment. To facilitate human-robot transfer, EgoSteer employs a unified action space of wrist poses and fingertip keypoints~\citep{cai2025n, fu2025metis}, coupled with training-time RTC~\citep{black2025training} to eliminate real-world execution pauses. Data-loading optimizations are further introduced to stabilize training at scale.

Empirically, through large-scale egocentric pre-training, real-robot post-training, and DAgger refinement, EgoSteer robustly follows free-form instructions across $45$ tasks with a $75\%$ average success rate. In cluttered scenes with multiple candidate tasks, it reliably executes commanded objectives while generalizing to novel skill compositions and unseen action semantics. Moreover, without explicit training, the policy exhibits notable instruction switching, immediately adapting mid-execution to abrupt command changes requiring drastically different motions. Together, these results confirm that EgoSteer genuinely grounds language guidance rather than memorizing spurious scene-task correlations. In comparative evaluations, EgoSteer consistently outperforms strong baselines such as $\pi_{0.5}$~\citep{intelligence2025pi05visionlanguageactionmodelopenworld} and Being-H0.5\citep{luo2026beingh05scalinghumancentricrobot}, with systematic ablations confirming the necessity of each system design. Additionally, the pre-trained manipulation priors also enable few-shot adaptation to challenging long-horizon tasks, such as box folding and cake unboxing, across two embodiments with $79\%$ average progress rate. Conversely, our from-scratch variant, $\pi_{0.5}$, and sample-efficient baselines including Diffusion Policy (DP)~\citep{chi2025diffusion} and IMLE~\citep{rana2025imle} struggle to make substantial progress, underscoring both task difficulty and the efficacy of our curated priors.
 To summarize, our contributions are five-fold:
\begin{itemize}[leftmargin=*, noitemsep, topsep=0pt, parsep=0pt, partopsep=0pt]
    \item \textbf{EgoSmith}, an egocentric data pipeline that curates a $9.6$K-hour fully-annotated pre-training corpus across 12 datasets, achieving a $8.3\times$ throughput and better accuracy over prior SOTA.
    \item \textbf{A unified robot stack} for teleoperation, inference, and seamless human-in-the-loop correction, that collects 187 hours of diverse data across 193 dexterous tasks.
    \item \textbf{EgoSteer}, a world-model-enhanced VLA operating on a morphology-aligned action space.
    \item \textbf{Extensive real-robot evaluations} demonstrating that EgoSteer robustly performs free-form language-following across 45 tasks and few-shot adapts to complex long-horizon tasks.
    \item \textbf{Open-source release} of our complete system, curated datasets, and model checkpoints.
\end{itemize}
\vspace{-8pt}
\section{Related Work}
\vspace{-8pt}
\label{sec:related-work}
\paragraph{Generalist robot manipulation policies.}
Building generalist policies has long been a core pursuit in embodied intelligence. Early work often adopts hierarchical architectures, decomposing the vision-language-to-action mapping through intermediate action representations~\citep{zhong2025surveyvisionlanguageactionmodelsaction}, such as trajectories~\citep{gu2024rttrajectory} and affordances~\citep{pmlr-v229-huang23b}, to leverage foundation model capabilities. However, these methods are typically constrained by the expressiveness of intermediate representations and compounding errors across modules, restricting their effectiveness to narrow scenarios. Subsequent research has shifted toward scaling diverse datasets for end-to-end training. Focusing on \textit{parallel-jaw grippers}, works such as $\pi_{0.5}$~\citep{intelligence2025pi05visionlanguageactionmodelopenworld}, $\pi_{0.7}$~\citep{pi07}, LingBot-VLA~\citep{lingbot-vla}, and G0.5~\citep{liu2026g0} scale tens of thousands of hours of robot data, while RDT-2~\citep{liu2026rdt2} leverages $10\text{K}$ hours of UMI~\citep{chi2024universal} demonstrations, achieving multi-task instruction following. In contrast, steerable manipulation on \textit{dexterous hands} remains largely unexplored, primarily due to the severe scarcity of open-source real-robot data and prohibitive data collection costs. To address this bottleneck, EgoSteer leverages egocentric human data for scalable pre-training and enables data-efficient post-training with fewer than $200$ hours of robot data, demonstrating steerability on dexterous hands.

Architecturally, a mainstream paradigm for end-to-end robot policies is vision-language-action (VLA) models, encompassing autoregressive formulations~\citep{kim2024openvlaopensourcevisionlanguageactionmodel,pertsch2025fast}, flow-matching experts~\citep{black2026pi0visionlanguageactionflowmodel}, and hybrid designs that combine both~\citep{driess2026knowledge}. More recently, world action models (WAMs) integrate action policies with future video or image prediction. Representative approaches include joint visual-action prediction~\citep{ye2026worldactionmodelszeroshot}, action generation conditioned on visual forecasts~\citep{li2026causal,zhang2026native, hu2026bagelvla}, decoupled visual and action prediction~\citep{yuan2026fastwam}, and unified models that simultaneously optimize diverse world modeling objectives~\citep{bi2026motus,agarwal2026cosmos,lyu2026lda}. However, policies reliant on visual generation typically suffer from heavy computational overhead and high inference latency, while unified models face challenges in balancing multiple learning objectives. To circumvent these limitations, EgoSteer adopts a lightweight paradigm by incorporating a world model objective into a flow-based VLA backbone: during training, it predicts action-conditioned future DINO features~\citep{simeoni2025dinov3} to endow the policy with foresight over physical outcomes, thereby enhancing instruction-following fidelity and manipulation precision; during inference, it generates only actions while discarding the visual prediction branch, ensuring minimal computational overhead and high-frequency closed-loop execution.
\vspace{-8pt}
\paragraph{Scaling with egocentric human data.}
Egocentric human videos offer a diverse and scalable data source for learning manipulation priors. Although numerous datasets have been released~\citep{hoque2025egodex, punamiya2026egoverse, li2026egolive}, their annotation comprehensiveness varies widely, with many containing only unannotated monocular RGB streams. While recent pipelines~\citep{li2025scalable, luo2026beingh05scalinghumancentricrobot} attempt to recover missing control modalities, they remain closed-source and constrained by throughput and reconstruction fidelity. To this end, our open-source EgoSmith pipeline improves upon HaWoR to achieve higher reconstruction precision alongside a $8.3\times$ throughput speedup, with rigorous multi-stage quality filtering. In downstream policy learning, existing works have verified that egocentric human data can improve robot performance~\citep{yang2025egovlalearningvisionlanguageactionmodels, fu2025metis, li2026egowam}, reduce real-robot data requirements~\citep{luo2025beingh0visionlanguageactionpretraininglargescale}, facilitate cross-embodiment transfer~\citep{luo2026beingh05scalinghumancentricrobot, zhang2026unidex}, exhibit scaling behaviors~\citep{zheng2026egoscalescalingdexterousmanipulation}, or serve as co-training data to enhance performance on grippers~\citep{pi07,wu2026foundation}. However, its effect on the steerability of dexterous hand models has been underexplored. We close this gap by proposing and open-sourcing a full-stack system that pre-trains a world-model-enhanced VLA on $9.6$K hours of curated egocentric data, post-trains it with $187$ hours of self-collected teleoperation data, and refines it with DAgger. The resulting model, EgoSteer, faithfully follows language instructions across over $40$ tasks, and we demonstrate through ablations that the key designs of our full-stack system are all beneficial to steerability.

\vspace{-8pt}
\section{EgoSmith: Curating Egocentric Videos into Dexterous Priors}
\vspace{-8pt}
\label{sec:egosmith}

\begin{figure}[t]
  \centering
  \includegraphics[width=\linewidth]{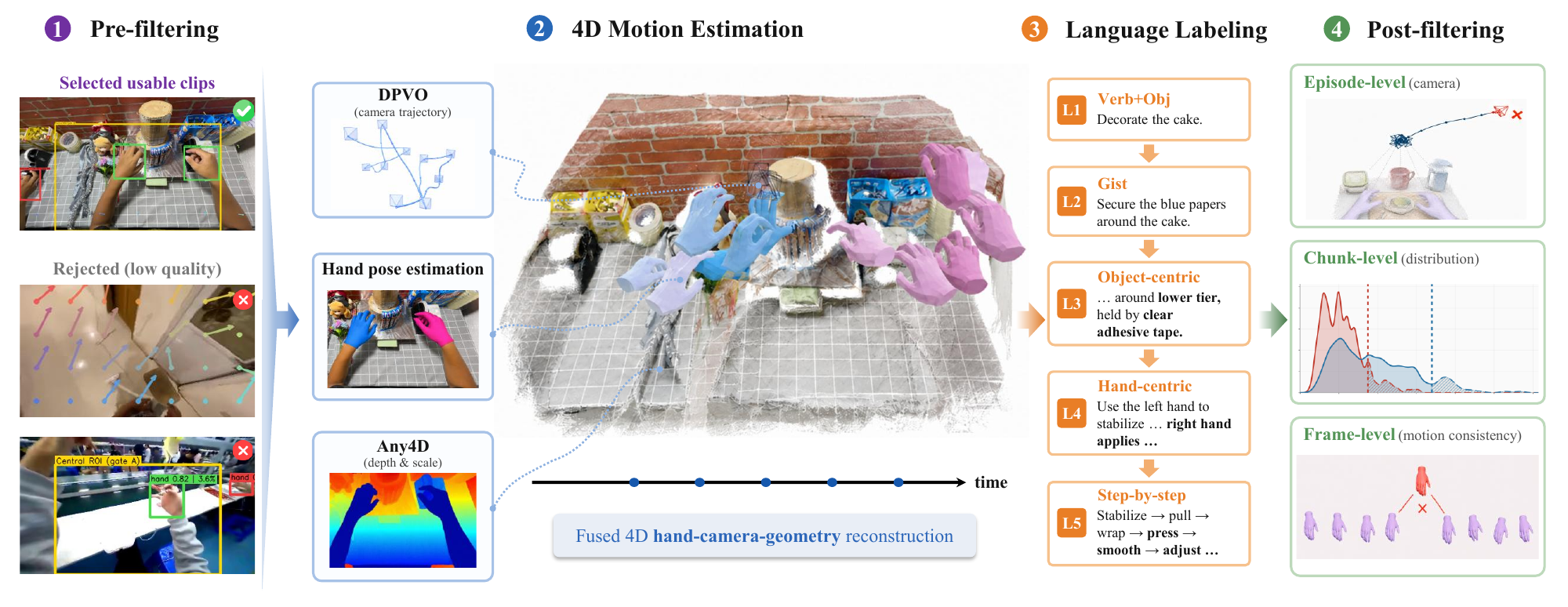}
  \caption{\textbf{Overview of EgoSmith}. Integrating pre-filtering, 4D motion estimation, language labeling, and post-filtering, EgoSmith efficiently curates in-the-wild egocentric videos into clean and annotated training samples.}
  \label{fig:egosmith}
  \vspace{-15pt}
\end{figure}

While egocentric data are rich in fine-grained interactions and highly scalable for general embodied learning, they are typically monocular RGB videos suffering from camera jitter, frequent occlusions, and a lack of annotations. This section presents \textbf{EgoSmith} (\Cref{fig:egosmith}), an efficient automated pipeline that transforms massive raw videos into fully-annotated training samples to enable effective learning. Full details can be found in \Cref{app:egosmith}.

To achieve raw data cleaning, labeling, and quality control, EgoSmith employs a four-stage pipeline. The \textbf{first} stage, \textbf{pre-filtering}, uses simple yet effective heuristics to discard locomotion segments, excessive head movements, and hand misidentifications that degrade downstream 4D motion estimation. Specifically, we filter out locomotion and abrupt head motion by detecting excessive camera motion via RANSAC-based~\citep{hartley2003multiple} sparse optical flow over a 128-point grid. We then eliminate frames with severe occlusions or bystander interference by applying geometric criteria to the hand counts, scales, and coordinates detected by YOLO~\citep{yolov3, potamias2025wilor}, preserving only clearly visible egocentric manipulations. 

\begin{wraptable}{r}{0.5\textwidth}
  \centering
  \vspace{-10pt}
  \caption{Benchmark results of 4D motion estimation.
  Errors are in mm; time is in seconds per episode.}
  \label{tab:egosmith}
  \vspace{-5pt}

  \small
  \begingroup
  \setlength{\tabcolsep}{4pt}
  \renewcommand{\arraystretch}{1.12}
  \resizebox{\linewidth}{!}{%
    \begin{tabular}{@{}lccccc@{}}
      \toprule
      Method
      & RPE $\downarrow$
      & ATE $\downarrow$
      & WA-MPJPE $\downarrow$
      & W-MPJPE $\downarrow$
      & Time $\downarrow$ \\
      \midrule
      HaWoR
      & 5.63 & 9.12 & 34.9 & 106.7 & 95.04 \\
      \textbf{EgoSmith} (Ours)
      & \textbf{2.60}
      & \textbf{7.57}
      & \textbf{26.7}
      & \textbf{88.2}
      & \textbf{11.45} \\
      \bottomrule
    \end{tabular}%
  }
  \endgroup

  \vspace{-10pt}
\end{wraptable}

Building upon the state-of-the-art method HaWoR, the \textbf{second} stage, \textbf{4D motion estimation}, reconstructs camera extrinsics, depth, and world-space hand trajectories. Since HaWoR lacks dense depth estimation and relies on DROID-SLAM~\citep{teed2021droid} for tracking, which is computationally expensive and prone to drift under rapid head movements and drastic scene changes, we propose a more robust and efficient scheme. We leverage DPVO~\citep{teed2023deep} for more stable, \textit{metric-free} camera tracking and keyframe depth estimation, and Any4D~\citep{karhade2025any4d} for frame-wise, \textit{metric-scale} depth prediction. Aligning their scale ratio recovers more accurate metric-scale camera trajectories, which we use to transform camera-frame hand motions into more consistent world-space trajectories. Beyond adopting DPVO, which is far more lightweight than DROID-SLAM, we incorporate batching and asynchronous I/O pipelining to further maximize processing throughput. Using NVIDIA A800 GPUs, we benchmark EgoSmith against HaWoR on a random subset across four datasets with ground-truth annotations: TACO~\citep{liu2024taco}, OakInk-v2~\citep{zhan2024oakink2}, H2O~\citep{kwon2021h2o}, and EgoVerse~\citep{punamiya2026egoverse}. This subset comprises $512$ episodes with an average duration of $15.26$ seconds at 30~FPS. As shown in \Cref{tab:egosmith}, EgoSmith outperforms HaWoR in both accuracy and efficiency. Lower RPE and ATE indicate that the DPVO-based camera trajectory estimation yields superior local consistency and structural robustness, while lower WA-MPJPE and W-MPJPE show that we effectively mitigate scale distortion and long-term drift, ensuring physically plausible world-space hand tracking over extended sequences. Moreover, EgoSmith reduces the per-episode runtime from $95.04$ to $11.45$ seconds, achieving an $8.3\times$ speedup to facilitate large-scale processing.

The \textbf{third} stage, \textbf{language labeling}, performs multi-granularity language annotation to support steerability~\citep{ChenW-RSS-26}. We first employ Qwen3.5-VL-Plus~\citep{qwen3.5} to filter out segments lacking meaningful hand-object manipulation. For the remaining clips, the model generates coarse-to-fine, five-level language instructions, ranging from concise verb-object phrases to detailed step-by-step procedures. This hierarchical annotation simultaneously provides task-level semantic grounding and action-level spatiotemporal grounding, enabling downstream models to learn and respond to instructions across varying levels of abstraction. 

The \textbf{fourth} stage, \textbf{post-filtering}, applies complementary quality checks at the episode, chunk, and frame levels. At the episode level, we compute the mean camera translation and rotation changes and reject episodes that are outliers with respect to their dataset-wide IQR distributions. At the chunk level, we transform wrist positions within a sliding window into the camera coordinate system of each anchor frame and express fingertip positions in the corresponding wrist coordinate systems. We reject episodes whose coordinate extrema fall outside dataset-derived IQR fences or predefined physical bounds. At the frame level, we measure consecutive-frame changes in camera translation and rotation, wrist translation and rotation, and fingertip translation. An episode is rejected if any of these changes exceeds physically plausible limits, thereby removing abrupt reconstruction jumps. Together, these checks remove unreliable episodes caused by discontinuous motion estimates, implausible metric scales, and camera-tracking drift.

With EgoSmith, we curate a large-scale egocentric pre-training corpus across 12 raw datasets~\citep{liu2024taco,kwon2021h2o,zhan2024oakink2,wang2023holoassist,garcia2018first,banerjee2025hot3d,damen2018scaling, grauman2022ego4d, punamiya2026egoverse,hoque2025egodex,buildaiegocentric10k2025,buildaiegocentric100k2025}. To filter out highly repetitive videos, we subsample Egocentric-10K~\citep{buildaiegocentric10k2025} and Egocentric-100K~\citep{buildaiegocentric100k2025}. For Ego4D~\citep{grauman2022ego4d} and EPIC-KITCHENS~\citep{damen2018scaling}, we apply EgoSmith to their respective VITRA~\citep{li2025scalable} subsets. Ultimately, this pipeline yields a fully-annotated egocentric dataset comprising $9.60\text{K}$ hours, $2.29\text{M}$ episodes, and $1.04\text{B}$ frames.

\vspace{-8pt}
\section{A Robot Stack for Teleoperation and DAgger Post-Training}
\vspace{-8pt}

\label{sec:robot-stack}

\begin{figure}[t]
  \centering
  \includegraphics[width=\linewidth]{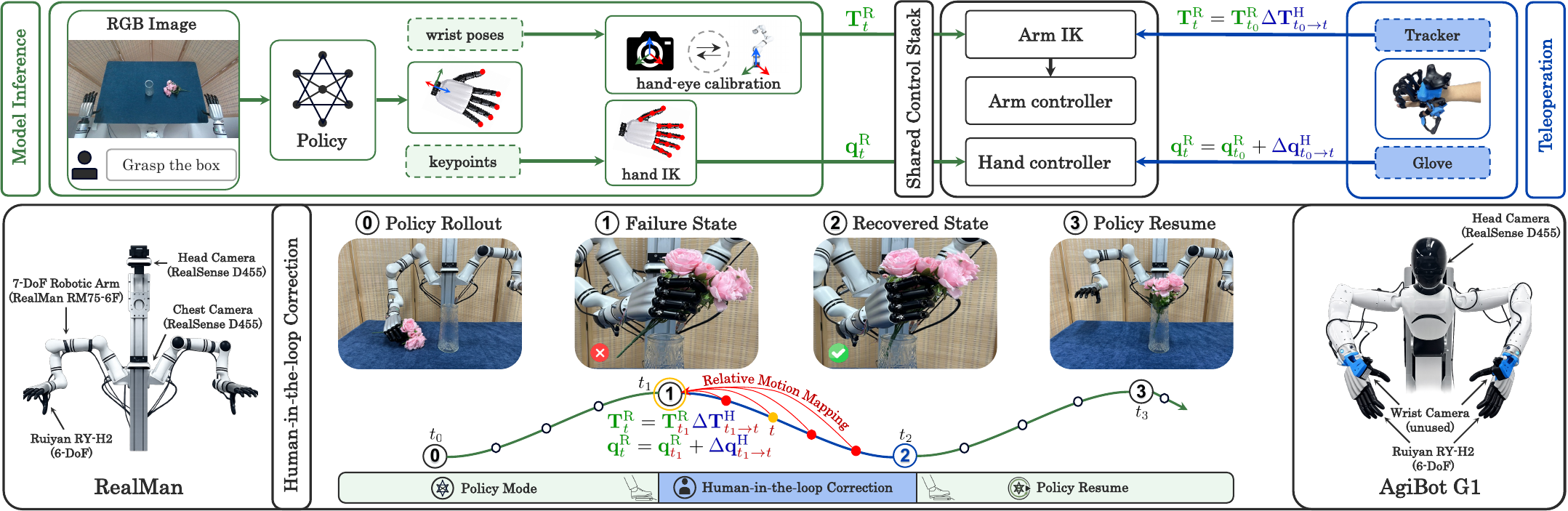}
  \vspace{-10pt}
\caption{\textbf{Overview of the Robot Stack}. It unifiedly supports teleoperation, policy inference, and human-in-the-loop correction. A relative motion mapping scheme is employed to facilitate seamless transitions during interventions, and the bottom row illustrates the two robotic embodiments utilized in our experiments.}
\label{fig:robot-stack}
  \vspace{-20pt}
\end{figure}

While EgoSmith's curated egocentric data provides rich manipulation priors, direct transfer to real robots is hindered by the embodiment gap across vision and control, necessitating real-robot teleoperation to ground these priors onto the target embodiment. This section presents the unified \textbf{Robot Stack} (\Cref{fig:robot-stack}), which shares low-level control and dynamics to simultaneously support teleoperation, policy inference, and human-in-the-loop correction. Full details can be found in \Cref{app:robot-stack}.

For teleoperation, a pair of PsiBot SynGlove-Air gloves and Vive Trackers capture the operator's $SE(3)$ wrist poses and hand joint angles, which respectively drive two robotic arms through inverse kinematics (IK) computed via \texttt{mink}~\citep{Zakka_Mink_Python_inverse_2026} and two 6-DoF robotic hands via joint mapping. During policy inference, the policy publishes wrist pose trajectories in the camera frame and hand keypoints in the wrist frame. These actions share the same arm and hand FK/IK and control nodes with teleoperation, ensuring identical execution dynamics across training and inference.

The primary challenge in enabling human-in-the-loop intervention is preventing sudden state jumps at the handover boundary to ensure a smooth transition. To address this, we propose a \textit{relative motion mapping} scheme. When the operator signals intervention by pressing a foot pedal at step $t$, the system records the robot end-effector poses $\mathbf{T}^{\text{R}, i}_t \in SE(3)$, human wrist poses $\mathbf{T}^{\text{H}, i}_t \in SE(3)$, robot hand joint states $\mathbf{q}^{\text{R}, i}_t \in \mathbb{R}^{6}$, and glove states $\mathbf{q}^{\text{H}, i}_t \in \mathbb{R}^{6}$ for each arm/hand index $i \in \{1, 2\}$. Subsequently, at any $t' \ge t$, the operator's relative motions, $\Delta\mathbf{T}_{t\rightarrow t'}^{\text{H},i} = (\mathbf{T}^{\text{H}, i}_t)^{-1} \mathbf{T}^{\text{H}, i}_{t'}$, $\Delta\mathbf{q}^{\text{H}, i}_{t\rightarrow t'} = \mathbf{q}^{\text{H}, i}_{t'} - \mathbf{q}^{\text{H}, i}_t$, are mapped to the robot, computing the commanded end-effector poses $\mathbf{T}^{\text{R}, i}_{t'}$ and hand joint states $\mathbf{q}^{\text{R}, i}_{t'}$ as $\mathbf{T}^{\text{R}, i}_{t'} = \mathbf{T}^{\text{R}, i}_t\Delta\mathbf{T}_{t\rightarrow t'}^{\text{H}, i}$ and $\mathbf{q}^{\text{R}, i}_{t'} = \mathbf{q}^{\text{R}, i}_t + \Delta\mathbf{q}^{\text{H}, i}_{t\rightarrow t'}$.
This formulation allows the operator to smoothly take over control by simply mimicking the robot's motion, enabling efficient collection of corrective demonstrations. 

\begin{figure}[t]
  \centering
  \includegraphics[width=\linewidth]{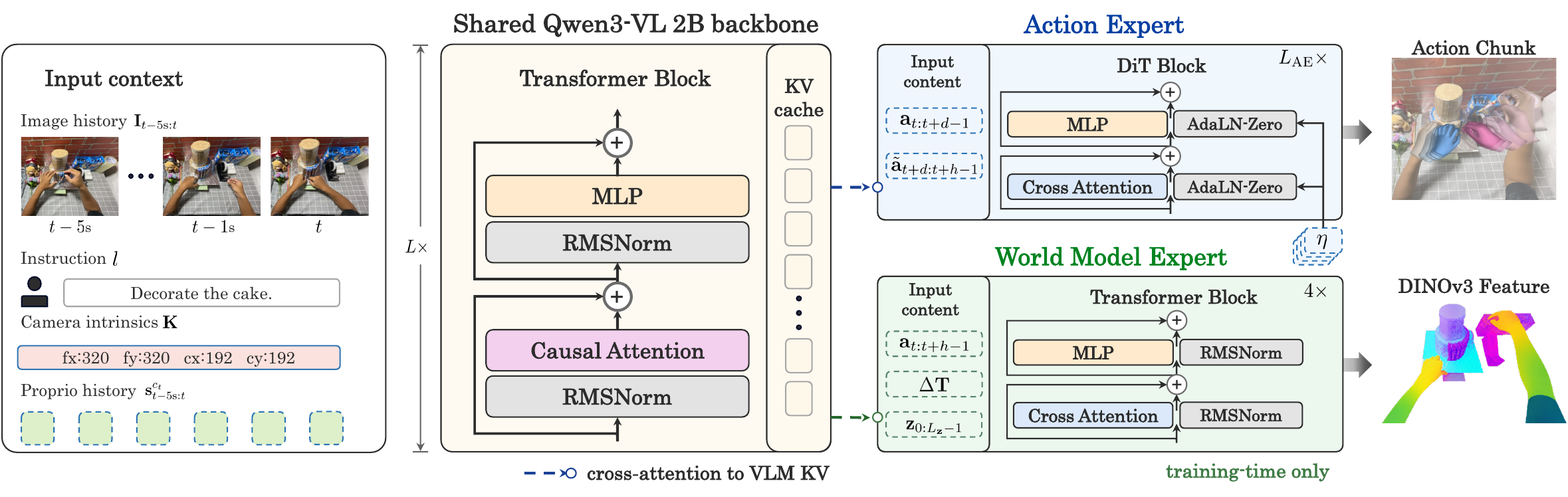}
\caption{\textbf{Overview of EgoSteer}, a world-model-enhanced VLA model for steerable dexterity. A shared Qwen3-VL backbone extracts KV cache representations from multi-modal inputs. The action expert jointly attends to itself and the backbone to generate action chunks via flow-matching, integrating training-time RTC to eliminate execution pauses. The training-only world model expert predicts future DINOv3 features to improve action accuracy with zero inference overhead.}
\label{fig:model-arch}
\vspace{-15pt}
\end{figure}

With unified robot stack, we collect a $187$-hour dataset across $193$ tabletop tasks spanning seven categories: Pick-and-Place(PnP)-Easy/Medium/Hard, Non-prehensile, Reorient, Bimanual, and Contact-rich. These comprise $56$ common tasks covering most core manipulation primitives, alongside $137$ long-tail tasks to facilitate human-to-robot transfer and grounding. Multi-level language annotations are generated using Qwen3-VL-Flash~\citep{Qwen3-VL}, followed by manual verification. To cover diverse primitives and establish modal alignment, data collection follows a free-form protocol where environments are cluttered, and tablecloths, object instances, and initial configurations are randomized without pre-defined trajectories. This emphasis on natural, human-like execution yields substantial rollout diversity, fostering policy robustness and multi-task generalization.
\vspace{-8pt}
\section{EgoSteer: A World-Model-Enhanced VLA for Steerable Dexterity}
\vspace{-8pt}
\label{sec:egosteer}

To effectively learn language-guided manipulation from human data, teleoperation, and corrections, we propose \textbf{EgoSteer}, a flow-based VLA model enhanced by a world-model objective, shown in \Cref{fig:model-arch}. To ensure robust vision-language understanding while modeling multimodal continuous actions, EgoSteer pairs a Qwen3-VL 2B backbone~\citep{Qwen3-VL} with a DiT-based~\citep{peebles2023scalable} action expert, which jointly attends to itself and backbone to generate action chunks via flow-matching~\citep{black2026pi0visionlanguageactionflowmodel}. To facilitate human-robot transfer, we design a unified data format and state-action space across both domains. An episode $\tau$ of length $N$ is represented as $\tau = \{ l, \mathbf{K}, (\mathbf{I}_t, \mathbf{D}_t, \mathbf{T}^{w2c}_t, \mathbf{s}^w_t, \mathbf{a}^w_t )^{N-1}_{t=0} \}$, where $l$ is the instruction, $\mathbf{K} \in \mathbb{R}^{3\times3}$ is the camera intrinsics, $t$ is the timestep, $\mathbf{I}_t \in \mathbb{R}^{H \times W \times 3}$ and $\mathbf{D}_t \in \mathbb{R}^{H \times W \times 1}$ are the RGB and depth images, $\mathbf{T}^{w2c}_t \in SE(3)$ is the world-to-camera extrinsics, and bimanual world-frame states and actions $\mathbf{s}^w_t, \mathbf{a}^w_t \in \mathbb{R}^{48}$ comprise the 3D wrist translation, 6D wrist rotation, and 15D fingertip keypoints of both hands. Beyond this unified action space, we align the robot's wrist coordinate frame to match human anatomical proportions, minimizing the domain gap to facilitate knowledge transfer (see \Cref{app:align-human-robot} for details). Since depth is unused in our model, the training sample at each timestep $t$ becomes $\{ l, \mathbf{K}, \mathbf{I}_{t-k+1:t}, \mathbf{s}^{c_t}_{t-k+1:t}, \mathbf{a}^{c_t}_{t:t+h-1} \}$, where $k$ and $h$ denote the history and prediction lengths, and $\mathbf{s}^{c_t}_{t-k+1:t}$ is the state history transformed into the current camera frame $c_t$ via $\mathbf{T}^{w2c}_t$. The relative action chunk $\mathbf{a}^{c_t}_{t:t+h-1}$ is computed in $c_t$ relative to $\mathbf{s}^{c_t}_t$, where wrist motions are relative $SE(3)$ transforms and finger movements are coordinate displacements. For simplicity, we omit the $c_t$ superscript for $\mathbf{s}$ and $\mathbf{a}$ hereafter. Furthermore, to avoid execution pauses during real-robot inference, we implement training-time Real-Time Chunking (RTC)~\citep{black2025training} in the action expert. Specifically, we feed a clean action prefix $\mathbf{a}_{\text{pre}} = \mathbf{a}_{t:t+d-1}$ of randomly sampled delay $d$ as ground truth and train the expert solely to denoise the subsequent actions $\tilde{\mathbf{a}}_{\text{suf}} = \tilde{\mathbf{a}}_{t+d:t+h-1}$. During deployment, the robot executes the reserved prefix $\mathbf{a}_{\text{pre}}$ during asynchronous VLA inference, transitioning seamlessly to the new chunk $\mathbf{a}_{\text{suf}}$ without execution gaps. Denote our model by $\pi$, we train it using Conditional Flow Matching (CFM)~\citep{lipmanflow} by regressing the linear velocity field of the target suffix $\mathbf{a}_{\text{suf}}$ conditioned on the context $\mathbf{C}_t = \{ l, \mathbf{K}, \mathbf{I}_{t-k+1:t}, \mathbf{s}_{t-k+1:t}, \mathbf{a}_{\text{pre}} \}$ with $\mathcal{L}_{\text{CFM}}(\pi) = \mathbb{E}_{t, \eta, \boldsymbol{\epsilon}} \left[ \| \pi(\tilde{\mathbf{a}}_{\text{suf}}, \eta, \mathbf{C}_t) - (\mathbf{a}_{\text{suf}} - \boldsymbol{\epsilon}) \|^2 \right]$, where $\eta \in [0,1]$, $\boldsymbol{\epsilon} \sim \mathcal{N}(\mathbf{0}, \mathbf{I})$, and $\tilde{\mathbf{a}}_{\text{suf}} = (1 - \eta)\boldsymbol{\epsilon} + \eta\mathbf{a}_{\text{suf}}$. To expand the effective batch size and improve loss gradient, we sample four random $\eta$ per sample. 

While VLA excels at vision-language understanding, lacking future imagination limits its action generation accuracy~\citep{ye2026worldactionmodelszeroshot}. To address this, we introduce a world-model expert to predict action-induced future DINOv3 features. The expert takes the ground-truth $\mathbf{a}_{t:t+h-1}$, the relative camera motion $\Delta\mathbf{T} = \mathbf{T}^{w2c}_t(\mathbf{T}^{w2c}_{t+h-1})^{-1}$, and learnable query tokens $\mathbf{z}_{0:L_\mathbf{z}-1}$ of length $L_\mathbf{z}$ as inputs to jointly attend to themselves and the backbone. The upsampled output of $\mathbf{z}$ is supervised against future frame $\mathbf{I}_{t+h-1}$ DINOv3 features via regression loss, which provides a more direct and stable supervision signal than generative loss. For robot setups with an additional chest camera, the backbone inputs both cameras' image histories and intrinsics, while the expert receives their relative motions and regresses both future DINOv3 features by adding distinct camera embeddings to $\mathbf{z}$. To focus optimization on the backbone's representation, this module comprises only four Transformer layers attending to the backbone layers at regular intervals, guiding the gradient to primarily shape the backbone. Crucially, the expert is discarded during inference, avoiding computational overhead.

\begin{figure}[t]
  \centering
  \vspace{-15pt}
  \includegraphics[width=\linewidth]{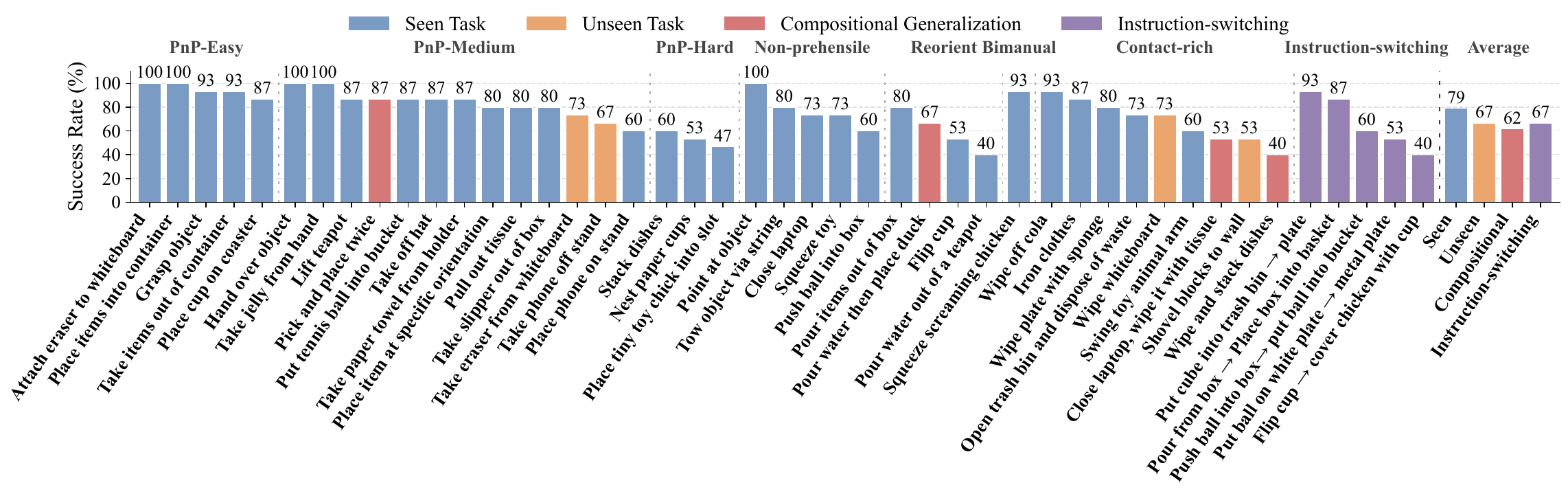}
\caption{Steerable manipulation performance of EgoSteer across 45 tasks spanning 8 categories. It robustly follows free-form language instructions to achieve an overall success rate of 75\%, demonstrating generalization.}
\label{fig:steerable-manipulation}
  \vspace{-15pt}
\end{figure}

Since our training samples encompass sparse multi-frame visual histories over extended temporal windows, naive random-access data loading imposes a prohibitive I/O burden. To address this, our infrastructure implements \textit{randomized sequential streaming} using WebDataset shards or LeRobot v3 episodes as streaming units. This scheme drastically reduces I/O pressure through sequential reading, while preserving training randomness via shuffle buffers, random sample dropping, and randomized unit selection. Additionally, we incorporate Hybrid Sharded Data Parallel (HSDP)~\citep{zhao2023pytorch}, mixed-precision training, kernel fusion, and FlexAttention~\citep{dong2024flex} to maximize compute throughput. Collectively, these optimizations ensure stable and highly efficient training at scale. Full details of EgoSteer and its training infrastructure can be found in \Cref{app:egosteer}.

\vspace{-8pt}
\section{Experiments}
\vspace{-8pt}
\label{sec:experiments}

We conduct extensive experiments to answer five core research questions:
\textbf{Q1}. How well does EgoSteer follow free-form instructions to complete various tasks? 
\textbf{Q2}. Does DAgger efficiently and effectively improve performance? 
\textbf{Q3}. How does the pre-training scale affect downstream performance, and how does EgoSteer compare with other VLA baselines? 
\textbf{Q4}. Are egocentric pre-training data quality, the world-model objective, and training-time RTC essential to strong performance? 
\textbf{Q5}. Can large-scale egocentric pre-training enable few-shot adaptation to complex long-horizon tasks? 

To thoroughly and rigorously investigate these questions, our evaluations utilize multiple EgoSteer checkpoints across various training configurations and stages. To prevent confusion, we outline their naming convention below. 
EgoSteer \textit{generalist} models follow a three-stage training pipeline:
\begin{enumerate}[label=\textbf{\arabic*.}, nosep, leftmargin=*]
    \item \textbf{Pre-training (PT)} on the egocentric dataset using a single head camera at $384 \times 384$ resolution.
    \item \textbf{Fine-tuning (FT)} on the robot dataset using dual head and chest cameras at $640 \times 480$ resolution.
    \item \textbf{DAgger refinement (DG)} under the identical dual-camera setup.
\end{enumerate}
In contrast, EgoSteer \textit{few-shot single-task} models require only two stages:
\begin{enumerate}[label=\textbf{\arabic*.}, nosep, leftmargin=*]
    \item \textbf{Pre-training (PT)} on the same single-camera egocentric data.
    \item \textbf{Few-shot adaptation (FS)} on a few robot demonstrations with the same single-camera setup.
\end{enumerate}
Accordingly, we designate our checkpoints as \textbf{EgoSteer-$n$K-PT/FT/DG/FS}, where $n\text{K}$ represents the hours of egocentric pre-training data (denoting a random subset of the full dataset when $0 < n < 9.6$, with $n = 0$ indicating no pre-training), while PT, FT, DG, and FS indicate that the model is trained up to the pre-training, fine-tuning, DAgger, and few-shot stages, respectively. EgoSteer-$n$K-FT models consistently employ the full $187$ hours of real-robot data for fine-tuning. Only the EgoSteer-$9.6$K-FT model undergoes DAgger refinement; specifically, three DAgger iterations are conducted, collecting $3.7$K trajectories across $56$ tasks, yielding $8.3$ hours of correction data to refine the policy. The resulting model is denoted as \textbf{EgoSteer-$9.6$K-DG}, which is also referred to simply as \textbf{EgoSteer}. Full experimental details and training configurations are provided in \Cref{app:experiments}.

\vspace{-8pt}
\subsection{Steerable Multi-Task Manipulation and Generalization}
\vspace{-8pt}
\label{subsec:steerable}
\textbf{Setup.} To evaluate the steerability of EgoSteer, we benchmark it across 45 tasks: $32$ seen, $4$ compositional generalization, $4$ unseen, and $5$ instruction-switching. \textit{Seen} tasks share action semantics with the fine-tuning dataset; \textit{compositional} tasks require recombining seen primitives into novel sequences; \textit{unseen} tasks introduce novel action semantics absent from fine-tuning; and \textit{instruction-switching} tasks assess the ability to respond to a new instruction during execution~\citep{chen2026resteer}. 
To ensure challenging evaluation conditions, tasks are performed in randomly cluttered scenes with multiple actionable distractors and commanded via unconstrained, template-free language. 
We evaluate each task over $15$ randomized trials, reporting average success rates in \Cref{fig:steerable-manipulation}.

\textbf{Results.} As shown in \Cref{fig:steerable-manipulation}, EgoSteer achieves 80+\% success on 24 tasks and an overall average of $75\%$. 
Crucially, across cluttered layouts, the policy strictly adheres to language instructions specifying target objects, hand assignments, and fine-grained motions, even when manipulating flat or small objects. 
Under identical initial scenes, EgoSteer faithfully carries out distinct tasks according to different instructions (\Cref{fig:steerability}). 
When prompted mid-execution with an abrupt command, EgoSteer can immediately respond and fulfill the new goal (\Cref{fig:resteer}). 
Furthermore, EgoSteer exhibits robust failure recovery, executing multiple retries if a step fails. 
It also achieves average success rates of $62\%$ on compositional generalization and $67\%$ on unseen tasks, confirming that our full-stack framework effectively harnesses rich priors from egocentric data to unlock robust, generalizable multi-task instruction following with fewer than 200 hours of real-robot demonstrations.

\vspace{-8pt}
\subsection{Efficacy of DAgger Post-Training}
\vspace{-8pt}
\label{subsec:dagger}

\textbf{Setup.} To evaluate the efficacy of DAgger, we compare EgoSteer-$9.6$K-DG and EgoSteer-$9.6$K-FT on four failure-prone, dexterous seen tasks, such as ``\textit{place phone on stand}''. For each task, 
$10$ evaluation trials are conducted using the same settings as in \Cref{subsec:steerable}.

\textbf{Results.} As shown in \Cref{tab:dagger}, after DAgger iterations totaling $8.3$ hours, the average success rate increases from $22.5\%$ to $62.5\%$. This efficacy stems from the targeted collection of corrective demonstrations addressing deployment failures, achieving a performance leap with minimal data. The refined policy not only exhibits robust failure recovery but also adaptively adjusts actions at critical manipulation bottlenecks. Crucially, these recovery and adjustment capabilities generalize to novel tasks, substantially improving the overall robustness of the DAgger-trained policy.

\begin{table*}[t]
\caption{Various experiments to validate the significance of system components and compare with baselines.}
\vspace{-5pt}
\label{tab:all-results}
\centering
\setlength{\tabcolsep}{1.5pt}
\renewcommand{\arraystretch}{1.08}

\newsavebox{\EgoTableA}
\sbox{\EgoTableA}{\normalsize
\begin{tabular}{@{}lc@{}}
    \toprule
    Method & Avg. \\
    \midrule
    EgoSteer-$9.6$K-FT & 22.5\% \\
    EgoSteer-$9.6$K-DG & \textbf{62.5\%} \\
    \bottomrule
\end{tabular}}

\newsavebox{\EgoTableB}
\sbox{\EgoTableB}{\normalsize
\begin{tabular}{@{}lc@{}}
    \toprule
    Method & Avg. \\
    \midrule
    $\pi_{0.5}$ & 22\% \\
    Being-H0.5 & 39\% \\
    EgoSteer-$9.6$K-FT & \textbf{74\%} \\
    \bottomrule
\end{tabular}}

\newsavebox{\EgoTableC}
\sbox{\EgoTableC}{\normalsize
\begin{tabular}{@{}lc@{}}
    \toprule
    Method & Avg. \\
    \midrule
    No WM-objective & 31\% \\
    No training-RTC & 39\% \\
    Noisy data & 33\% \\
    EgoSteer-$1$K-FT & \textbf{44\%} \\
    \bottomrule
\end{tabular}}

\newsavebox{\EgoTableD}
\newcommand{\EgoWMFile}{figures/wm_paper_cup.pdf}
\sbox{\EgoTableD}{%
    \resizebox{!}{\dimexpr\ht\EgoTableC+\dp\EgoTableC\relax}{%
        \vbox{%
            \offinterlineskip
            \hbox{\includegraphics[
                width=143.462bp,
                trim=0bp 52.64326bp 0bp 0bp,
                clip
            ]{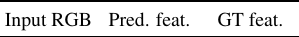}}%
            \hbox{%
                \parbox[c][13.56bp][c]{143.462bp}{%
                    \centering
                    \fontsize{7.5bp}{9bp}\selectfont
                    Instruction: \textbf{Stack a cup pyramid}\par
                }%
            }%
            \hbox{\includegraphics[
                width=143.462bp,
                trim=0bp 9.221bp 0bp 26.1946bp,
                clip
            ]{\EgoWMFile}}%
            \hbox{\includegraphics[
                width=143.462bp,
                trim=0bp 0bp 0bp 66.057bp,
                clip
            ]{\EgoWMFile}}%
        }%
    }%
}

\resizebox{\linewidth}{!}{%
\begin{minipage}{\dimexpr\wd\EgoTableA+\wd\EgoTableB+\wd\EgoTableC+\wd\EgoTableD+30pt\relax}
\captionsetup[subtable]{justification=centering}

\begin{subtable}[t]{\wd\EgoTableA}
    \caption{DAgger ablation on four failure-prone tasks.}
    \label{tab:dagger}
    \centering
    \usebox{\EgoTableA}
\end{subtable}%
\hfill
\begin{subtable}[t]{\wd\EgoTableB}
    \caption{Baseline comparison.}
    \label{tab:baselines}
    \centering
    \usebox{\EgoTableB}
\end{subtable}%
\hfill
\begin{subtable}[t]{\wd\EgoTableC}
    \caption{Training ablation.}
    \label{tab:ablation}
    \centering
    \usebox{\EgoTableC}
\end{subtable}%
\hfill
\begin{subtable}[t]{\wd\EgoTableD}
    \caption{World-model visualization.}
    \label{tab:world-model}
    \centering
    \usebox{\EgoTableD}
\end{subtable}%

\end{minipage}%
}
\vspace{-12pt}
\end{table*}

\vspace{-8pt}
\subsection{Scaling of Pre-Training and Baseline Comparisons}
\vspace{-8pt}
\label{subsec:scaling-and-baseline}

\textbf{Setup.} To investigate the scaling behavior of egocentric data, we compare the pre-training loss curves and robot performance of EgoSteer-$0$\,/\,$3$\,/\,$6$\,/\,$9.6$K-FT across $10$ tasks. Additionally, the baselines $\pi_{0.5}$ and Being-H0.5 are post-trained on our robot data and evaluated across $10$ simpler tasks.

\begin{wrapfigure}{r}{0.33\textwidth}
    \centering
    \includegraphics[width=0.31\textwidth]{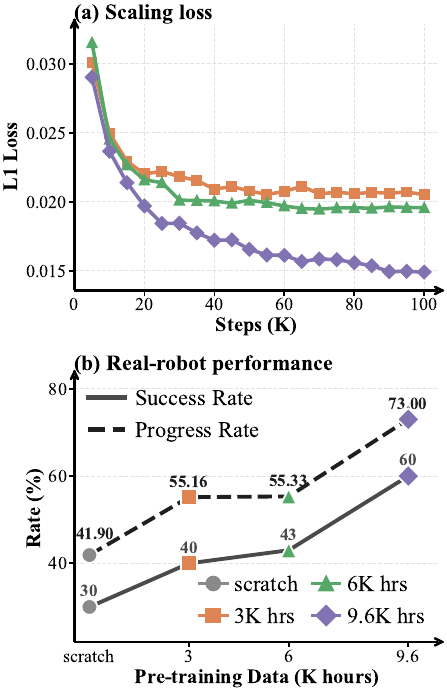}
    \caption{Scaling behavior of pre-training loss and downstream real-robot post-training performance.}
    \label{fig:scaling_law}
    \vspace{-1.8em}
\end{wrapfigure}

\textbf{Results.} As shown by the pre-training loss curves in \Cref{fig:scaling_law}a and the real-robot success and progress rates in \Cref{fig:scaling_law}b, scaling pre-training data drives training loss to lower convergence values while improving real-world execution performance. With expanding pre-training data, the policy exhibits the emergence of failure recovery, enhanced instruction-following, and improved action accuracy, indicating that the model successfully acquires physical common sense for error adjustment and language-guided manipulation priors from increasingly larger datasets curated by EgoSmith.

The comparison of EgoSteer-$9.6$K-FT with the baselines is presented in \Cref{tab:baselines}, where EgoSteer consistently outperforms both. Specifically, both baselines lack dedicated mechanisms for knowledge transfer to new dexterous hands, thereby limiting the efficacy of pre-training. Consequently, although they can handle basic PnP actions, they exhibit weak instruction-following, limited generalization, and imprecise execution. These performance gains highlight the advantage of our unified, full-stack system.

\vspace{-8pt}
\subsection{Ablation Studies}
\vspace{-8pt}
\label{subsec:ablation}
\textbf{Setup.} We compare EgoSteer-$1$K-FT with three ablated variants across $10$ seen tasks: first, \textit{No WM-objective}, which omits the world-model expert during pre- and post-training; second, \textit{No training-RTC}, which disables training-RTC during training and inference; and third, \textit{Noisy data}, which utilizes noisy egocentric data unfiltered by EgoSmith pre- and post-filtering.

\textbf{Results.} As shown in \Cref{tab:ablation}, removing any core component leads to a performance drop, validating the necessity of each module. Specifically, the \textit{No WM-objective} variant exhibits a marked reduction in fine-grained manipulation accuracy, confirming that enhancing the backbone's action imagination is critical for precise action generation. As visualized in \Cref{tab:world-model}, predictions from EgoSteer's world model expert align closely with the ground truth. The \textit{No training-RTC} variant introduces severe action pauses and disrupts execution dynamics, causing contact-rich tasks to fail entirely due to continuous jitter. Finally, the \textit{Noisy data} variant fails to converge effectively, leading to severe degradation in both instruction-following and manipulation precision.

\begin{figure*}[!t]
    \centering
    \includegraphics[width=\textwidth]{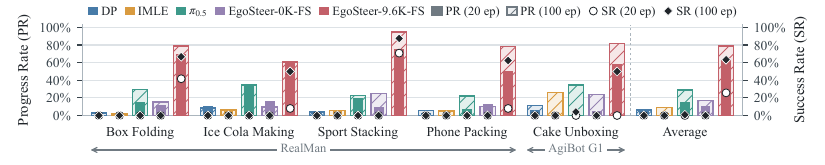}
    \captionsetup{skip=1pt}
    \caption{Few-shot performance across five challenging tasks on two embodiments. Ep stands for episodes.}
    \label{fig:single-task}
    \vspace{-15pt}
\end{figure*}

\vspace{-8pt}
\subsection{Few-Shot Adaptation to Complex Long-Horizon Tasks}
\vspace{-8pt}
\label{subsec:single-task}
\textbf{Setup.} To investigate whether manipulation priors from egocentric pre-training promote sample-efficient robot learning, we few-shot adapt EgoSteer-9.6K-PT to five challenging long-horizon tasks, such as ``\textit{box folding}'', across two embodiments using $20$ and $100$ demonstrations. These tasks comprise roughly ten fine-grained sub-steps and span approximately one minute. We benchmark EgoSteer-9.6K-FS against our non-pretrained ablation, $\pi_{0.5}$, and strong imitation learning baselines, namely DP and IMLE, over $24$ real-world trials per task. Evaluation results and task setups are reported in \Cref{fig:single-task} and illustrated in \Cref{fig:single-task-settings}, respectively.

\textbf{Results.} As shown in \Cref{fig:single-task}, EgoSteer-$9.6$K-FS consistently outperforms all baselines. Due to the extreme task complexity, all baselines make negligible progress even with $100$ demonstrations. In contrast, using merely $20$ demonstrations, EgoSteer-$9.6$K-FS achieves substantial progress and notable success rates; increasing to $100$ demonstrations further yields marked improvements across both metrics. These results confirm that our $9.6$K-hour pre-training imparts rich dexterous priors, enabling effective few-shot transfer to novel embodiments and complex manipulation tasks.


\vspace{-8pt}
\section{Conclusion}
\vspace{-8pt}
\label{sec:conclusion}

This paper presents and open-sources a full-stack system for steerable dexterous manipulation that scales egocentric pre-training and enables data-efficient real-robot post-training. It integrates EgoSmith, an efficient egocentric data pipeline; a unified robot stack for teleoperation, inference, and correction; and EgoSteer, a world-model-enhanced VLA with a morphology-aligned action space. EgoSteer demonstrates robust steerability across 45 diverse tasks, exhibiting dexterity, failure recovery, and generalization, while achieving few-shot adaptation to long-horizon tasks on two embodiments. Baseline comparisons and ablations confirm the criticality of system components.



\clearpage
\section*{Acknowledgments}
We sincerely thank Bo Liu and Shaoyang Guo for their generous help. We also extend our gratitude to our colleagues at PsiBot, including but not limited to Xiaojie Chai, Jianxin Du, Lin Huang, Ruochong Li, Haoyi Su, Tang Li, Yunlong Wang, Hongze Yu, Chaoyang Liu, and Hui Zhang, for their valuable support and helpful discussions.



\bibliography{iclr2027_conference}
\bibliographystyle{iclr2027_conference}  

\clearpage

\appendix
\startcontents[appendix]

{\noindent\huge\bfseries Appendix\par}
\vspace{0.8em}
{\noindent\Large\bfseries Table of Contents\par}
\vspace{3pt}
\hrule height 0.8pt
\vspace{-0.4em}

{
\hypersetup{linkcolor=black}
\printcontents[appendix]{}{1}{\setcounter{tocdepth}{3}}
}

\nobreak\vspace{3pt}
\hrule height 0.8pt
\clearpage

\section{Details of EgoSmith}
\label[appendix]{app:egosmith}
\vspace{-4pt}

This section presents the implementation details of EgoSmith (\Cref{app:egosmith-implementation}) and provides detailed statistics of the curated $9.6\text{K}$-hour pre-training corpus derived from 12 egocentric human video datasets (\Cref{app:curated-dataset-statistics}).

\vspace{-4pt}
\subsection{Implementation Details}
\label[appendix]{app:egosmith-implementation}
\vspace{-4pt}

\subsubsection{Pre-Filtering Heuristics}
\label[appendix]{app:pre-filtering}
\vspace{-4pt}

The pre-filtering stage employs frame-wise heuristics to rapidly discard low-quality segments, such as those containing locomotion, excessive head movement, hand absence, occlusion, or others' hand interference. These five scenarios are handled by two specialized gates: a camera gate for motion-related issues, and a hand gate for visibility anomalies. A contiguous segment is pruned only if it contains at least three consecutive invalid frames; isolated failures are retained as they exert negligible impact on subsequent reconstruction.

The camera gate estimates ego-motion using sparse optical flow. For each frame, a 128-point grid is tracked back 15 frames earlier via pyramidal Lucas–Kanade~\citep{bouguet2001pyramidal}. We fit a similarity transform using RANSAC~\citep{hartley2003multiple}; the frame passes this gate if the translation of the estimated transform is within $10\%$ of the image's longer dimension.

The hand gate uses YOLO~\citep{yolov3, potamias2025wilor} to detect hands, retaining a bounding box as valid only if it satisfies three criteria: (a) confidence $\ge 0.30$ to reject false positives; (b) area within $[2\%, 50\%]$ of the image, where the $50\%$ upper bound excludes hands abnormally close to the lens, and the $2\%$ lower bound is calibrated on Egocentric-10K/100K~\citep{buildaiegocentric10k2025, buildaiegocentric100k2025} subsets, where we manually labeled small bounding boxes to find a threshold that filters out most other people's hands while retaining the operator's; and (c) spatially intersecting with the lower-central region (normalized $[0.075, 0.925]$ horizontally, $[0.075, 1.0]$ vertically). The gate requires $\ge 2$ valid detections per frame, which naturally filters out hand absence, occlusion, or cases where only other people's hands are present, while preserving clear bimanual manipulation. 

Together, the two gates yield video segments characterized by stable camera motion and clearly visible bimanual interactions.

\vspace{-4pt}
\subsubsection{4D Motion Estimation}
\label[appendix]{app:4d-motion-estimation}
\vspace{-4pt}

In stage 2, we reconstruct hand motions in a unified, metric world-space coordinate system. Within this pipeline, we adopt the ViT module from HaWoR as an off-the-shelf camera-frame hand reconstructor. Specifically, hands are detected and cropped across frames, and the ViT processes these inputs in temporal windows to regress the frame-wise MANO~\citep{MANO} pose parameters $\boldsymbol{\theta}_t\in\mathbb{R}^{51}$, shape parameters $\boldsymbol{\beta}_t\in\mathbb{R}^{10}$, and the camera-relative root translation $\mathbf{t}_t \in \mathbb{R}^3$. Although this camera-space hand reconstruction is highly reliable, placing these reconstructions in world space with accurate physical dimensions requires a robust, metric camera trajectory. This is a primary limitation of the original HaWoR pipeline, as its dependency on DROID-SLAM can accumulate drift under egocentric conditions with rapid head movements or textureless environments. To address this, we design a robust pipeline to recover a metric, temporally consistent camera trajectory in world space, onto which the hand reconstructions are subsequently mapped.

We replace DROID-SLAM with DPVO~\citep{teed2023deep} to estimate the camera trajectory. DPVO is more robust in long-range egocentric scenarios and incurs much lower computational cost, outputting up-to-scale camera poses $\hat{\mathbf{T}}_t = (\mathbf{R}_t, \hat{\mathbf{p}}_t) \in SE(3)$, where the hat notation $\hat{\cdot}$ denotes up-to-scale quantities. We then anchor the trajectory to the physical scale using metric depth estimates from Any4D~\citep{karhade2025any4d} as a reference. To ensure temporal coherence across the entire sequence, we perform a cross-chunk alignment on the local Any4D depth windows, yielding a temporally consistent metric depth sequence.

To recover the metric scale factor $s$ for DPVO trajectory,
we compute the median ratio of the aligned Any4D depth $\mathbf{D}^{\mathrm{Any4D}}_t$ to the DPVO depth $\hat{\mathbf{D}}^{\mathrm{DPVO}}_t$ over the background pixels across all keyframes:
\begin{equation}
s = \operatorname{median}_{t, (u,v) \in \mathcal{B}_t} \frac{\mathbf{D}^{\mathrm{Any4D}}_t(u,v)}{\hat{{\mathbf{D}}}^{\mathrm{DPVO}}_t(u,v)},
\end{equation} 
where $\mathcal{B}_t$ is the valid background region in frame $t$, defined by excluding the hand regions projected from our reconstructed 3D hand mesh. We then calibrate the entire trajectory via $\mathbf{p}_t = s \hat{\mathbf{p}}_t$, yielding a metric camera trajectory.

Given the metric camera poses, we transform the camera-space hand vertices and joints, denoted generally as 3D coordinates $\mathbf{x}^{\mathrm{cam}}_t \in \mathbb{R}^3$, into world space via $\mathbf{x}^{\mathrm{world}}_t = \mathbf{R}_t^\top(\mathbf{x}^{\mathrm{cam}}_t - \mathbf{p}_t)$. The output for each video segment consists of the world-space bimanual states and actions, frame-wise camera intrinsics and extrinsics $(\mathbf{K}_t, \mathbf{T}_t)$, MANO parameters, and the Any4D metric scene depth.

Concerning efficiency, stage 2 represents the primary computational bottleneck of our entire pipeline, with most of its overhead arising from the two core components of 4D motion estimation: camera trajectory estimation and hand motion reconstruction. For camera tracking, replacing the dense DROID-SLAM with the substantially more lightweight DPVO significantly reduces the associated computational cost. Hand reconstruction, meanwhile, involves repeated ViT inference over temporal windows, for which the original HaWoR processes only a single 16-frame window at a time. We therefore group multiple windows into a single batch for parallel forward passes and asynchronously overlap CPU-based frame decoding and cropping with GPU-based model inference to minimize GPU idling. By improving the efficiency of both major components, these optimizations collectively accelerate the complete 4D motion estimation stage, achieves an overall speedup of approximately 8.3× compared to HaWoR.

For completeness, we provide detailed definitions of the four metrics used in Table~\ref{tab:egosmith} to jointly evaluate camera-trajectory and world-space hand-motion accuracy (all errors are reported in mm):
\begin{itemize}[leftmargin=*, noitemsep, topsep=0pt]
\item \textbf{Relative Pose Error (RPE)}: Measures local camera tracking drift over a fixed temporal interval without global trajectory alignment, making it sensitive to local motion errors and metric-scale inaccuracies.
\item \textbf{Absolute Trajectory Error (ATE)}: Measures global camera trajectory error after aligning the estimation to the ground truth with a single $\mathrm{Sim}(3)$ transformation, thereby evaluating trajectory structure and long-term drift while factoring out global scale and coordinate-frame differences.
\item \textbf{World-Aligned Mean Per Joint Position Error (WA-MPJPE)}: Measures world-space hand joint error after aligning the predicted and ground-truth hand trajectories with a single $\mathrm{Sim}(3)$ transformation over each 100-frame segment. Unlike per-frame alignment, this preserves temporal trajectory consistency and penalizes relative drift within the segment.
\item \textbf{World Mean Per Joint Position Error (W-MPJPE)}: Measures world-space hand joint error after rigidly aligning only the first frame of each 100-frame segment with an $\mathrm{SE}(3)$ transformation. It therefore preserves the predicted metric scale and accumulates subsequent scale, translation, rotation, and temporal drift errors, providing the strictest evaluation of world-space hand tracking.
\end{itemize}

\subsubsection{Language Labeling Prompt}
\label[appendix]{app:egosmith-language-labeling}

Below, we present the prompt template designed for Qwen3.5-VL-Plus~\citep{qwen3.5} to generate multi-granularity language annotations for egocentric human videos.

\begin{breakableprompt}{Egocentric Human Video Language Labeling Prompt}
# Video Annotation Task: General Egocentric Hand Action Description

## Objective
Provide a comprehensive 5-level language description for pre-cropped egocentric videos of human hand-object interaction. These videos may show daily activities, tool use, object handling, cooking, cleaning, repair, assembly, organization, inspection, or other practical manipulation tasks.

## Context
- **Perspective:** Ego-view focused on the hands and manipulated objects.
- **Purpose:** Training data for robotic systems.
- **Core Focus:** The visible hand action, the manipulated object, contact points, grasp type, spatial relationships, and the physical sequence needed to complete the task.

---

## Task Instructions

Analyze the video clip and perform the following steps:

### Step 1: Content Filtering (Hand-Action Status Check)
- **Mark "status": "Invalid"** if the video shows:
    - Walking, camera transition, or scene scanning with no meaningful hand-object interaction.
    - Passive observation with no active manipulation.
    - Hands resting, hanging, or only briefly entering the frame without acting on an object.
    - Non-manipulation activities such as talking, reading, waiting, or looking around.
    - A task where the main hand action cannot be identified because of severe occlusion, blur, or ambiguity.
- **Mark "status": "Valid"** if the video shows active hand-object manipulation, such as picking, placing, opening, closing, pouring, wiping, folding, pressing, turning, cutting, fastening, arranging, inserting, removing, scanning a barcode/card/object, or operating an object.
- If the main action is identifiable but partially occluded, mark `"status":"Valid"`.

### Step 2: Multi-Level Hand Action Instructions (For "Valid" only)

- **Level 1 (Verb + Object):** Core task. Max 5 words. Example: "Open the drawer."
- **Level 2 (Gist):** Concise summary of the hand action. Max 15 words.
- **Level 3 (Object-Centric):** Describe the manipulated object, relevant parts, state, and spatial features. Max 30 words.
- **Level 4 (Hand-Centric):** Specify left/right hand roles, grasp style, contact points, and coordination. Max 50 words.
- **Level 5 (Dense Sequence):** Step-by-step physical breakdown. Max 100 words.
    - Use spatial anchors such as "near the left edge," "above the bowl," "inside the slot," or "against the surface."
    - Describe motion trajectories such as "lift upward," "slide forward," "rotate clockwise," "press downward," or "pull toward the body."
    - Describe outcome state such as "fully seated," "opened," "aligned," "placed flat," "wiped clean," or "released."

---

## Strict Formatting & Quality Requirements

- **Verb-first imperative:** Start with an action verb. **NO subjects**.
- **Definite object references:** Use "the" for visible objects; avoid "a" or "an" in the generated instructions.
- **No transitional words:** Omit "then", "next", "afterwards".
- **Action Precision:** Use specific physical verbs such as "Grip," "Lift," "Place," "Open," "Close," "Pour," "Fold," "Wipe," "Press," "Turn," "Insert," "Remove," "Align," "Slide," "Scoop," "Cut," "Peel," "Tighten," or "Release" when possible.
- **Avoid over-specialization:** Do not infer hidden intent, brand names, object identities, or materials unless clearly visible.

---

## Output Format
Return a single JSON object. No markdown code fences. No extra text.
For `"status":"Invalid"`, return empty strings for all five levels.

{
  "status": "Valid/Invalid",
  "language_instructions": {
    "level1": "<verb and object>",
    "level2": "<concise summary>",
    "level3": "<object-focused description>",
    "level4": "<hand/object interaction details>",
    "level5": "<dense physical step-by-step>"
  }
}

\end{breakableprompt}

\subsubsection{Post-Filtering Criteria}
\label[appendix]{app:post-filtering}

In this stage, we perform quality control on the reconstructed outputs to filter out reconstruction anomalies and problematic segments, ensuring overall data quality. This evaluation is conducted from coarse to fine across three granularities: entire episodes, chunk windows, and adjacent frames.

At the episode level, we characterize overall camera motion using the mean frame-to-frame changes in the translation and rotation components of the camera extrinsics. Specifically, for each episode, we average the magnitudes of the consecutive-frame translation changes and rotation-matrix differences. We then compare these two episode-level motion statistics against their respective distributions over the episodes in the current dataset processing batch. Because typical camera motion varies across capture devices, scenes, and manipulation styles, we use dataset-dependent IQR fences rather than fixed thresholds. An episode is rejected if either statistic falls outside $[Q_1 - 2.5\,\mathrm{IQR},\; Q_3 + 2.5\,\mathrm{IQR}]$. This criterion targets episodes with atypical camera trajectories caused by tracking failures or unusually large head and body motion.

Chunk-level checks evaluate whether hands fall within physically reasonable spatial boundaries in a standardized egocentric coordinate frame. Directly comparing absolute hand coordinates is problematic, since they are coupled with camera and body movements. Specifically, within a sliding window spanning approximately the past 5 seconds and the future 30 frames, we transform all hand states and actions into the current camera frame. Under this canonical system, wrist positions are defined relative to the camera, and finger joints relative to the wrist. Within this coordinate system, we evaluate both distributional outliers and absolute physical limits. First, outliers in wrist translation/rotation and finger positions are identified using the same IQR criterion based on the respective dataset's distribution. Second, we enforce a universal physical ceiling of 1.5 meters on each coordinate axis for the hands, as a human hand cannot physically reach further than this distance from the head. If any sliding chunk window within an episode violates either the IQR outlier threshold or the 1.5-meter physical limit, the entire episode is discarded.

Frame-level checks identify sudden jumps between adjacent frames. We compute the frame-to-frame changes in camera translation and rotation, wrist translation and rotation, and finger translation. Unlike the previous levels, we do not rely on dataset-specific distributions here, as the physical speed of human hands and heads has a universal limit. Any movement exceeding this limit is attributed to reconstruction artifacts rather than valid motion. We therefore apply fixed physical thresholds: camera translation $\le 0.20\,\text{m/frame}$, wrist and finger translation $\le 0.30\,\text{m/frame}$, camera rotation $\le 28^\circ/\text{frame}$, and wrist rotation $\le 41^\circ/\text{frame}$. An episode is discarded if any of its frames violate these thresholds.

Collectively, these three levels of checks filter out problematic segments caused by head tracking drift, inaccurate motion reconstruction, and motion discontinuities.

\subsection{Curated Dataset Statistics}
\label[appendix]{app:curated-dataset-statistics}

The final pre-training corpus is constructed by utilizing EgoSmith to process 12 raw egocentric datasets, ultimately yielding a standardized, $9.6\text{K}$-hour dataset. This curated corpus comprises world-space bimanual states and actions, frame-wise camera intrinsics and extrinsics, metric scene depth, and coarse-to-fine language annotations. In this section, we detail the scale, source composition, and semantic diversity of this dataset. The primary objective is not merely scaling up dataset volume; instead, the core advantage lies in data quality. Every sample is richly annotated and filtered through a rigorous quality-control pipeline, guaranteeing high-quality, modality-aligned manipulation knowledge to assist the model in learning steerable dexterous manipulation. The dataset further offers broad coverage of manipulation tasks, objects, and multi-granularity language descriptions.

\begin{figure}[tp]
  \centering
  \begin{minipage}{\linewidth}
    \centering
    \small
    \setlength{\tabcolsep}{6pt}
    \renewcommand{\arraystretch}{1.15}
    \resizebox{\linewidth}{!}{%
\begin{tabular}{lrrrcccc}
    \toprule
    \textbf{Dataset} & \textbf{Hours} & \textbf{Percentage (\%)} & \textbf{Episodes} & \textbf{Hand} & \textbf{Depth} & \textbf{Camera} & \textbf{Language} \\
    \midrule
    \rowcolor{zebra} \textbf{Egocentric-100K}~\citep{buildaiegocentric100k2025} & $8,049$ & $83.8$ & $1{,}795{,}731$ & \cmark & \cmark & \cmark & \cmark \\
    \textbf{EgoVerse}~\citep{punamiya2026egoverse} & $690$ & $7.2$ & $35{,}175$ & & \cmark & & \cmark \\
    \rowcolor{zebra} \textbf{EgoDex}~\citep{hoque2025egodex} & $370$ & $3.9$ & $147{,}588$ & & \cmark & & \cmark \\
    \textbf{Egocentric-10K}~\citep{buildaiegocentric10k2025} & $288$ & $3.0$ & $194{,}915$ & \cmark & \cmark & \cmark & \cmark \\
    \rowcolor{zebra} \textbf{Ego4D}~\citep{grauman2022ego4d} & $138$ & $1.4$ & $74{,}505$ & \cmark & \cmark & \cmark & \cmark \\
    \textbf{Epic-Kitchens}~\citep{damen2018scaling} & $49$ & $0.5$ & $26{,}454$ & \cmark & \cmark & \cmark & \cmark \\
    \rowcolor{zebra} \textbf{HoloAssist}~\citep{wang2023holoassist} & $11.5$ & $0.1$ & $11{,}426$ & & \cmark & & \cmark \\
    \textbf{HOT3D}~\citep{banerjee2025hot3d} & $4.5$ & $0.05$ & $1{,}105$ & & \cmark & & \\
    \rowcolor{zebra} \textbf{TACO}~\citep{liu2024taco} & $3.0$ & $0.03$ & $1{,}558$ & & & & \cmark \\
    \textbf{OakInk-v2}~\citep{zhan2024oakink2} & $1.7$ & $0.02$ & $891$ & & \cmark & & \\
    \rowcolor{zebra} \textbf{H2O}~\citep{kwon2021h2o} & $1.0$ & $0.01$ & $935$ & & & & \\
    \textbf{FPHA}~\citep{garcia2018first} & $0.5$ & $0.01$ & $578$ & \cmark & \cmark & & \cmark \\
    \midrule
    \textbf{Total} & $\mathbf{9,606}$ & $100$ & $\mathbf{2{,}290{,}861}$ & & & & \\
    \bottomrule
    \end{tabular}%
}
    \subcaption{Dataset composition and annotation sources}
    \label{tab:egosmith_composition}
  \end{minipage}

  \vspace{1.6em}

  \begin{minipage}[t]{0.48\linewidth}
    \centering
    \includegraphics[width=\linewidth]{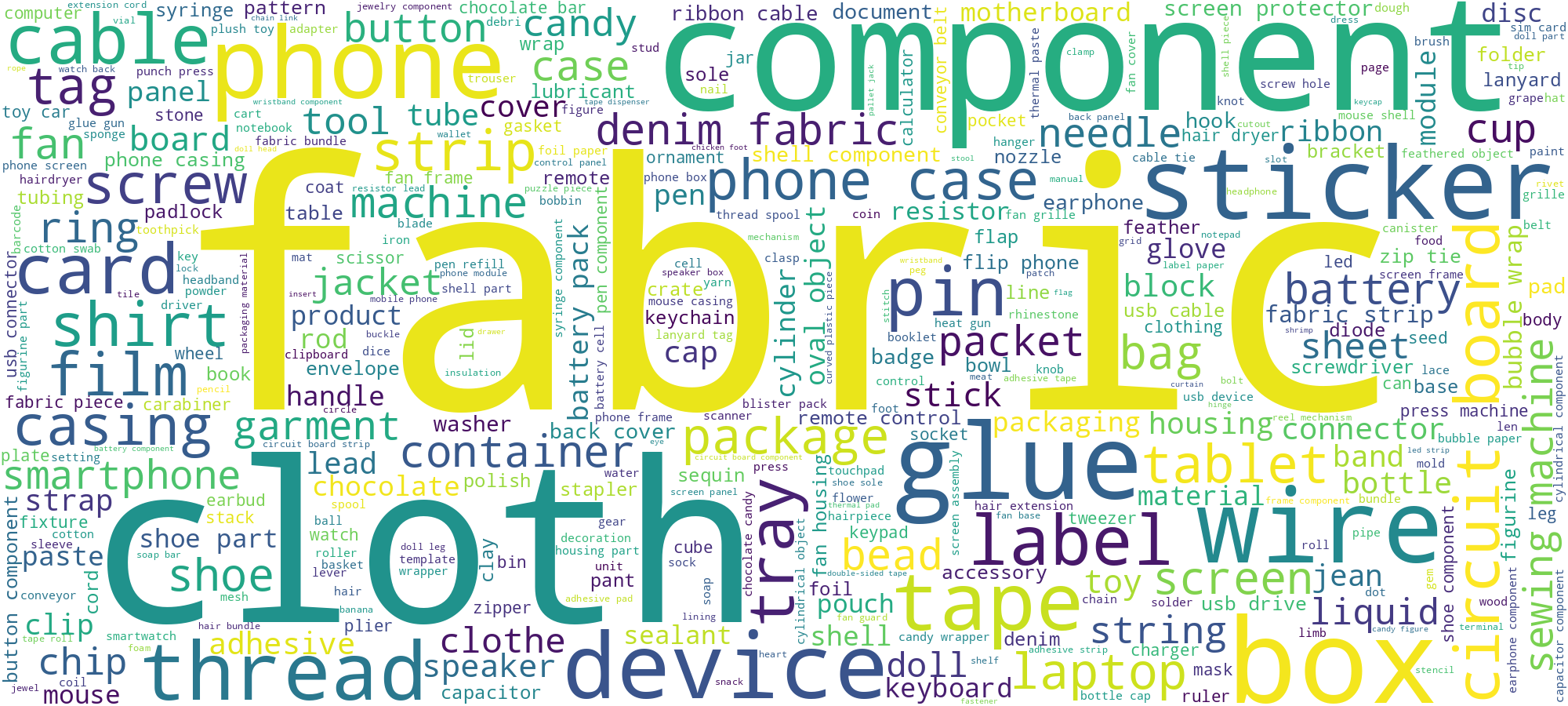}
    \subcaption{Word cloud of manipulated objects}
    \label{fig:egosmith_object}
  \end{minipage}\hfill
  \begin{minipage}[t]{0.48\linewidth}
    \centering
    \includegraphics[width=\linewidth]{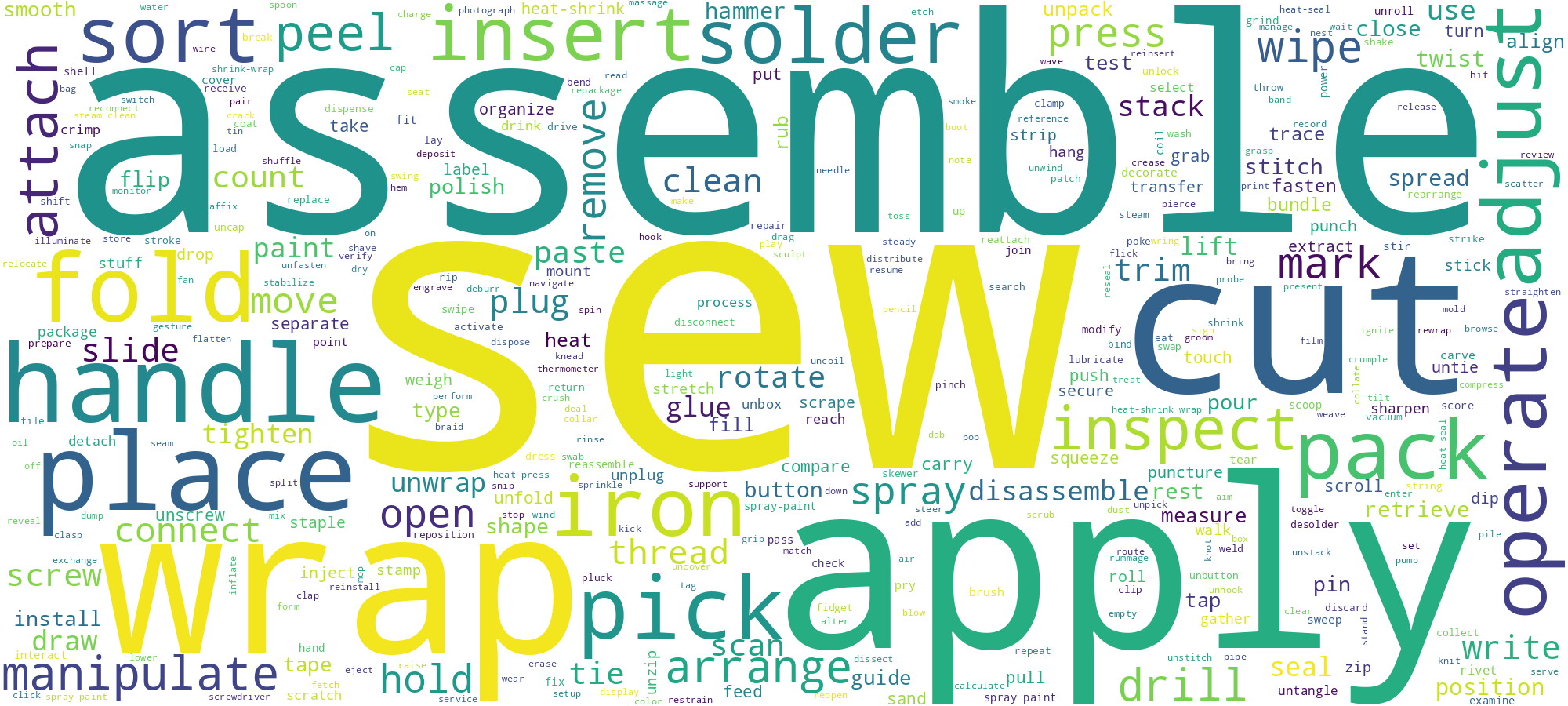}
    \subcaption{Word cloud of action verbs}
    \label{fig:egosmith_verb}
  \end{minipage}

  \vspace{1.6em}

  \begin{minipage}[t]{\linewidth}
    \centering
    \includegraphics[width=\linewidth]{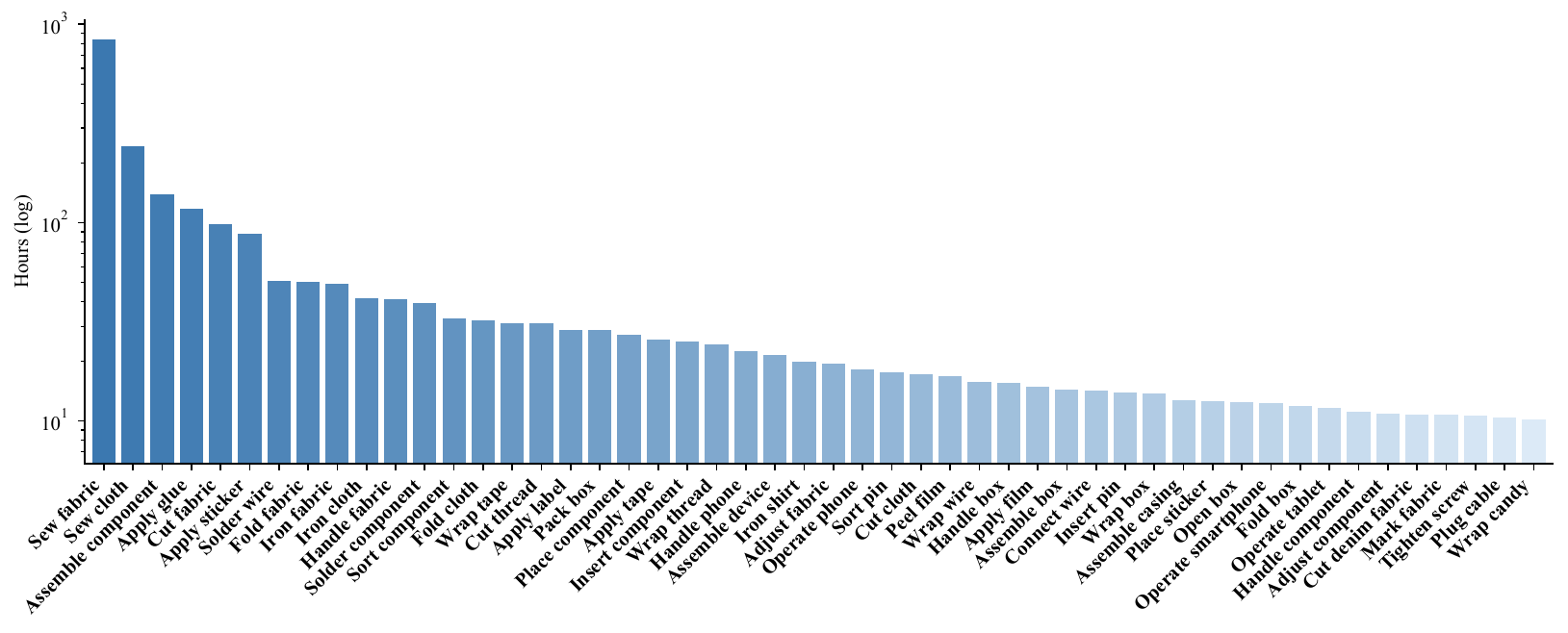}
    \subcaption{Top-$50$ ``verb\,+\,object'' atomic tasks}
    \label{fig:egosmith_top_vo}
  \end{minipage}
  \vspace{0.6em}

  \caption{\textbf{Curated dataset statistics.} (a) Source composition of the $12$ egocentric datasets, detailing duration, percentage, episode count, and per-annotation origin. Checkmarks denote annotations generated by our pipeline, while blank entries indicate natively provided annotations. For FPHA, the checkmark under \emph{Hand} specifically denotes the additionally annotated second hand. (b)--(d) Task and semantic diversity: word clouds of the manipulated objects and action verbs, and the most frequent verb--object atomic tasks, exhibiting long-tailed coverage over diverse manipulation skills.}
  \label{fig:egosmith_dataset_charts}
\end{figure}

\noindent\textbf{Scale and Source Composition.} \Cref{tab:egosmith_composition} presents the duration contribution and proportion of each source dataset, while also indicating whether each annotation is natively provided or reconstructed by our pipeline. The source datasets are highly complementary, covering a wide range of devices, scenes, and manipulation styles. Although a small number of large-scale collections dominate the total duration, the breadth of sources contributes substantial diversity.

\noindent\textbf{Task and Semantic Diversity.} To characterize the manipulation span of our dataset, we extract (verb, object) tuples from the L1 verb--object annotations and analyze the distributions of action verbs and manipulated objects. The dataset covers 8,969 distinct object nouns and 623 action verbs (\Cref{fig:egosmith_object,fig:egosmith_verb}). \Cref{fig:egosmith_top_vo} illustrates the top 50 most frequent ``verb\,+\,object'' atomic tasks. Common tasks concentrate on fundamental manipulation skills, aligning with the natural distribution of daily hand--object interactions. At the same time, the distribution exhibits a prominent long tail: a large number of low-frequency yet semantically diverse tasks and objects provides broad coverage, preventing the dataset from being biased toward a few dominant actions. This combination of common fundamental skills, long-tail task diversity, and multi-granularity language annotations provides downstream models with abundant training samples, broad task coverage, and rich dexterous manipulation knowledge. 

\clearpage
\section{Details of the Robot Stack}
\label[appendix]{app:robot-stack}

This section details the implementation of the Robot Stack (\Cref{app:robot-stack-implementation}), alongside the collection protocols and statistics of the $187$-hour real-robot dataset (\Cref{app:teleoperation}).

\subsection{Implementation Details}
\label[appendix]{app:robot-stack-implementation}

\subsubsection{Hardware Setup}
\label[appendix]{app:hardware-setup}

The two physical embodiments utilized in our work are illustrated in \Cref{fig:robot-stack}. The primary platform, referred to as the RealMan embodiment, consists of two 7-DoF RealMan RM75-6F robotic arms and two 6-DoF Ruiyan RY-H2 dexterous hands. It is equipped with one head-mounted and one chest-mounted Intel RealSense D455 camera, providing dual egocentric viewpoints to ensure a comprehensive visual field. The AgiBot G1 embodiment is adapted from an AgiBot G1 humanoid robot by replacing its default end-effectors with two 6-DoF Ruiyan RY-H2 hands. It features one head-mounted Intel RealSense D455 camera and two wrist cameras, the latter of which are unused. During experiments, the robot's neck, waist, and mobile base are kept fixed. Both platforms conduct tabletop manipulation on an operational table. The RealMan platform serves as the primary embodiment, on which the entire $187$-hour dataset was collected. All experiments are conducted on this setup, with the sole exception of the few-shot adaptation experiment performed on the AgiBot G1.

For both platforms, the human wrist tracker, arm kinematic solvers, and joint control ROS 2 nodes operate at $100$~Hz, while the glove, hand solvers, and control ROS 2 nodes run at $80$~Hz. This high-frequency loop ensures near-zero latency, enabling intuitive bimanual teleoperation and the collection of fine-grained manipulation demonstrations. The cameras capture frames at $30$~Hz. Data is recorded at these native frequencies and subsequently resampled to $30$~Hz for training.

\subsubsection{Aligning Robot Data with Egocentric Human Data}
\label[appendix]{app:align-human-robot}

Robot data collection aims to ground the human manipulation priors learned from large-scale pre-training onto the target physical embodiment. To preserve and transfer this pre-trained knowledge, minimizing the domain gap between robot and human data is essential. Because the robotic palm is slightly longer than a human palm, we translate the robot's wrist coordinate frame axially forward. This alignment ensures that the scale from the wrist to the fingertips matches human anatomical proportions. When converting raw robot data into training data, this coordinate transformation is applied alongside hand-eye calibration to map hand actions into the camera coordinate frame. Correspondingly, during real-time policy inference, the control stack executes the inverse transformation and hand-eye mapping to project predicted actions back to the robot's physical coordinate space.

\subsubsection{Hand-Eye Verification and Re-Calibration}

In practice, hand-eye calibration is prone to drift due to mechanical maintenance, wear, or collisions, with such discrepancies sometimes only identified \textit{post}-acquisition. To ensure data quality, we introduce an offline verification and automated recalibration pipeline. RGB-D images from the camera are projected into a 3D point cloud, and the robot's 3D mesh is rendered in the same space based on the hand-eye calibration, utilizing their spatial overlap for validation. Upon detecting misalignment, our pipeline runs FoundationPose~\citep{wen2024foundationpose} on the point cloud to estimate the robot hand's 6D pose based on its URDF. Through FK, the calibration matrix is back-calculated for each frame. Averaging these matrices across frames and removing outliers robustly recovers the calibration parameters, preventing data degradation.

\subsubsection{Language Labeling Prompt}

Below, we present the prompt template designed for Qwen3-VL-Flash~\citep{Qwen3-VL} to generate multi-granularity language annotations for teleoperated demonstrations.

\begin{breakableprompt}{Robot Data Language Labeling Prompt}
**CRITICAL VISUAL CONTEXT & PRIOR (READ CAREFULLY):**
You are observing two synchronized egocentric videos (Head View: top, Chest View: bottom) of an agent performing a manipulation task. This output will be used as language instructions for robotic training.
1. **The Agent's Hand:** The moving entity is the agent's bare end-effector. It generally has a grey base and black fingers.
2. **COLORED TIPS WARNING:** The tips/pads of the fingers often have GREEN, ORANGE, or RED tape/markers on them. THESE ARE PART OF THE FINGERS. They are NOT separate tools.
3. **EMPTY-HANDED PRIOR:** The agent is operating empty-handed. NEVER describe the agent as holding or using a 'green-tipped tool', 'hot knife', 'pliers', or any handheld instrument.
4. **ABSOLUTE GROUND TRUTH (TASK ALIGNMENT):** The specific task is **[{task_name}]**. This task name is your absolute ground truth for interpreting WHAT is being manipulated (Objects) and HOW it is being manipulated (Verbs). You MUST use the exact nouns implied by the task name.
5. **GRAMMAR:** Your description must be in **simple present tense**. Write in fluent English, avoid awkward phrasing.
                
**OBJECTIVE:**
Describe the agent's actions (focusing strictly on hand-object interactions) in **simple present tense** by integrating information from both views into a single, unified description at three levels of detail. **Since this is for robot training, you MUST completely ignore all task-irrelevant items.**
                
"**CONSTRAINTS:**
1. **Unified Description:** Provide ONE consolidated set of descriptions.
2. **Levels of Detail:**
   - **Level 1 (Gist):** A concise summary of the main action. Should always be a verb+noun phrase (i.e Ring a bell) (<20 words).
   - **Level 2 (Descriptive):** Main action + features and spatial layout of the **ACTIVELY MANIPULATED OBJECTS ONLY** (<40 words). **DO NOT list or describe any stationary background clutter.** In your description, list which hand is performing the action.
   - **Level 3 (Sequential):** The step-by-step temporal flow of key functional actions (<70 words). Include essential phases (reach -> manipulate -> release) but OMIT trivial micro-adjustments or hovering. List which hand is performing each action.
3. **Zero Subjects (Strict):** **Start every single sentence directly with a verb** (e.g., 'Reach for...', 'Grasp...'). DO NOT use subjects (e.g., 'The person', 'The robot', 'The hand', 'It').
4. **Strict Focus on Interaction (NO CLUTTER):** Focus ONLY on the objects being actively touched, moved, or interacted with (and their immediate targets/receptacles). **Completely IGNORE all irrelevant background items** (e.g., wipes, boxes, tubes, stands that are not part of the task). Never write phrases like 'other items remain unchanged' or 'in the background'.
5. **Vocabulary Restrictions:**
   - DO NOT output words like 'robot', 'mechanical arm', 'gripper', 'human', or 'finger'.
   - DO NOT output colors of the agent's hand/tips.
6. **Task Vocabulary (Verbs & Nouns):** Your choice of verbs AND target nouns must strictly align with the task **[{task_name}]**.
7. **Action Logic & Validation:** Focus on the actual state changes of the objects. Verify actual contact using both views.
8. **Object Disambiguation:** Use spatial descriptors (e.g., 'the topmost card') ONLY for task-relevant items to distinguish them from each other.
9. **Tense:** Use **Simple Present** tense (e.g., 'reach', 'grasp', 'slide').
10. **Spatial Description:** Use the camera frame as the reference. 
   - Use **'upper', 'middle', 'lower'** to describe distance of objects on table. For example, 'the upper left of the table' refers to the far side of the table, and 'the lower left' refers to the near side.
   - Use 'left' 'right' to describe horizontal relationships\n"
   - Use 'on', 'on top of', 'above', 'below' etc. to describe vertical relationships.

                
**OUTPUT FORMAT:**
1. [Level 1 Description]
2. [Level 2 Description]
3. [Level 3 Description]


**EXAMPLE (If Task Name is 'Draw_cards'):**
1. Draw playing cards.
2. Slide the top cards from a central deck with left hand to draw them to the lower part of the table.
3. Reach toward the central deck with left hand, press down on the topmost card, and slide it backward. Return to the deck, press on the next card, and slide it backward to complete the draw."
\end{breakableprompt}

\subsection{Teleoperation Data Collection}
\label[appendix]{app:teleoperation}

Although pre-training on egocentric human videos equips the model with rich dexterous manipulation priors, direct physical deployment is prevented by the embodiment gap across visual appearance, dynamics, and kinematics. Furthermore, because these pre-training labels are generated by automated depth and hand pose estimation models, their accuracy is constrained by current model capabilities, potentially introducing systematic biases into the pre-trained policy. Therefore, collecting real-robot teleoperation data aims to correct and ground these priors onto the target embodiment with high sample efficiency. Additionally, due to kinematic constraints on robotic degrees of freedom, the robot cannot perfectly replicate all fine-grained human hand movements. To achieve steerable manipulation, we must collect a highly diverse range of free-form tasks to maximize the coverage of basic manipulation primitives, guiding the model to efficiently transfer cross-domain knowledge and fostering robust compositional generalization and multi-task instruction following.

Based on these considerations, within the kinematic limits of RealMan's degrees of freedom, we design $193$ semantically distinct dexterous manipulation tasks. For each task, approximately $300$ randomized, diverse demonstration trajectories, totaling around $1$ hour, are collected under cluttered scenarios, constructing a high-quality real-robot dataset of $187$ hours and $55\text{K}$ trajectories.

These $193$ tasks are classified into two major categories:
\begin{itemize}[leftmargin=*, noitemsep, topsep=0pt]
    \item \textbf{Common Tasks} ($56$ tasks): Everyday manipulations that are readily achievable within the current robot hardware configuration and sensor limits, yielding high teleoperation success rates.
    \item \textbf{Long-Tail Tasks} ($137$ tasks): Infrequent and physically challenging manipulations, such as contact-sensitive operations lacking tactile feedback, with lower collection success rates, primarily designed to ensure comprehensive semantic coverage of the dexterous manipulation space.
\end{itemize}

Based on motion and physical interactions, all tasks are further categorized into seven classes:
\begin{itemize}[leftmargin=*, noitemsep, topsep=0pt]
    \item \textbf{PnP-Easy}: Single-step tabletop pick-and-place where objects are easily graspable and the placement space is open.
    \item \textbf{PnP-Medium}: Non-planar or 3D spatial pick-and-place that requires precise operations related to containers, demanding higher control accuracy and spatial perception, such as ``\textit{put tennis ball into ball holder}''.
    \item \textbf{PnP-Hard}: Multi-step or high-precision pick-and-place sequences, such as ``\textit{stack paper cups}''.
    \item \textbf{Non-prehensile}: Non-grasping actions, including pushing, pulling, and pressing.
    \item \textbf{Reorient}: Actions involving rotation and reorientation, such as ``\textit{pour water}'' or ``\textit{flip paper cups}''.
    \item \textbf{Bimanual}: Bimanual tasks requiring high synchronization and spatial coordination between arms and hands, such as ``\textit{plug cable into charger}''.
    \item \textbf{Contact-rich}: Operations involving frequent and complex physical contact interactions, requiring physical understanding, such as ``\textit{wipe whiteboard}''.
\end{itemize}

Throughout data collection, strict requirements are enforced on the randomness, diversity, and quality of each trajectory. Specifically, the tabletop features cluttered, unstructured scenes rather than pre-arranged, simplified environments, avoiding task identification from visual inputs alone. This design forces the model to deeply align language instructions with physical actions, encouraging it to learn general task semantics and execution goals rather than memorizing demonstration trajectories. Furthermore, given the highly randomized object configurations, operators are instructed to perform teleoperation in a natural, human-like manner. Consequently, different demonstrations of the same task exhibit substantial variations in execution trajectories, thereby covering a broader distribution of the mapping from human priors to the robot's action space.

\begin{figure}[p]
  \centering

  \begin{minipage}[t]{0.44\linewidth}
    \centering
    \includegraphics[width=\linewidth]{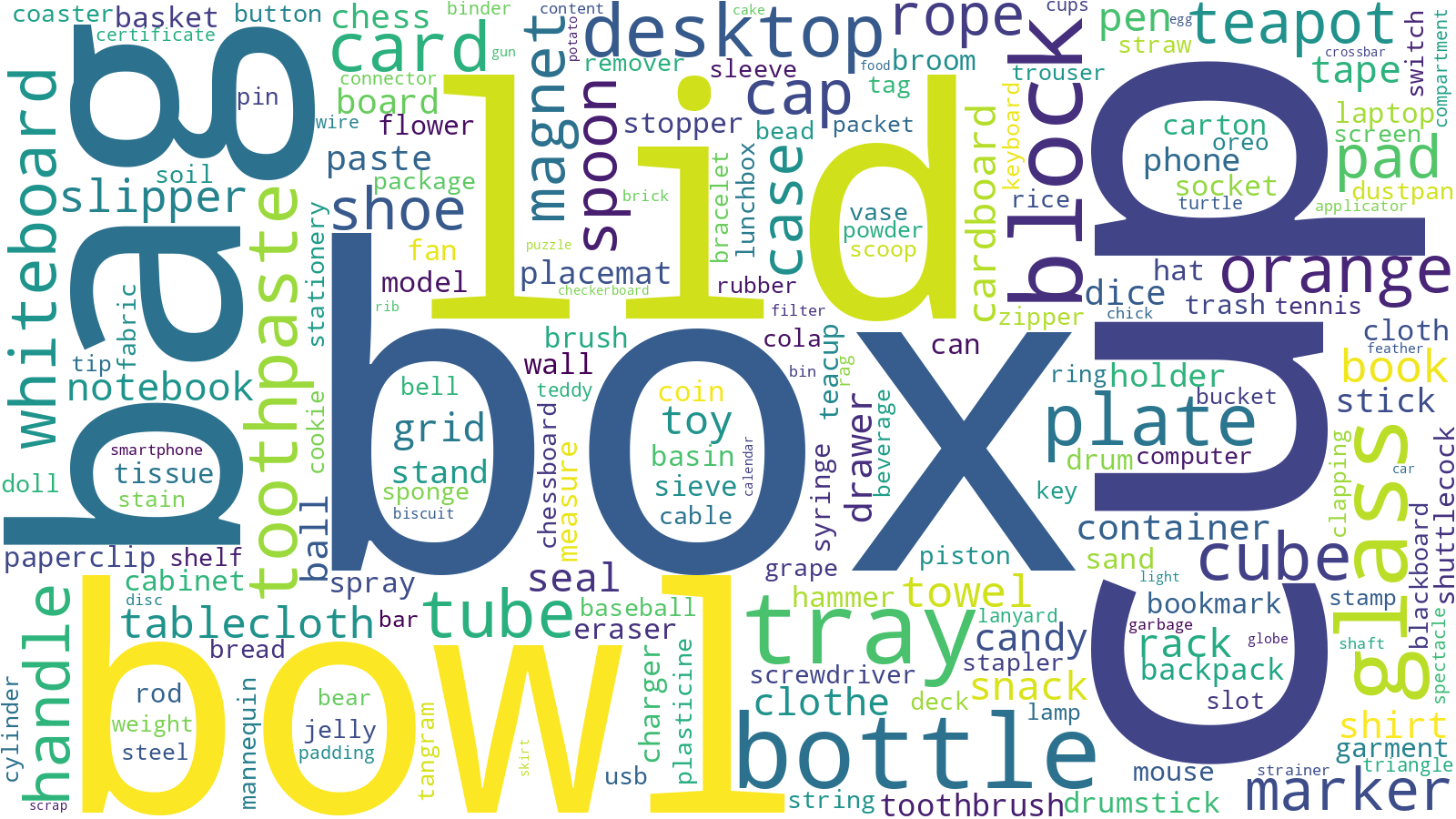}
    \subcaption{Word cloud of nouns in annotation}
    \label{fig:noun}
  \end{minipage}\hfill
  \begin{minipage}[t]{0.52\linewidth}
    \centering
    \includegraphics[width=\linewidth]{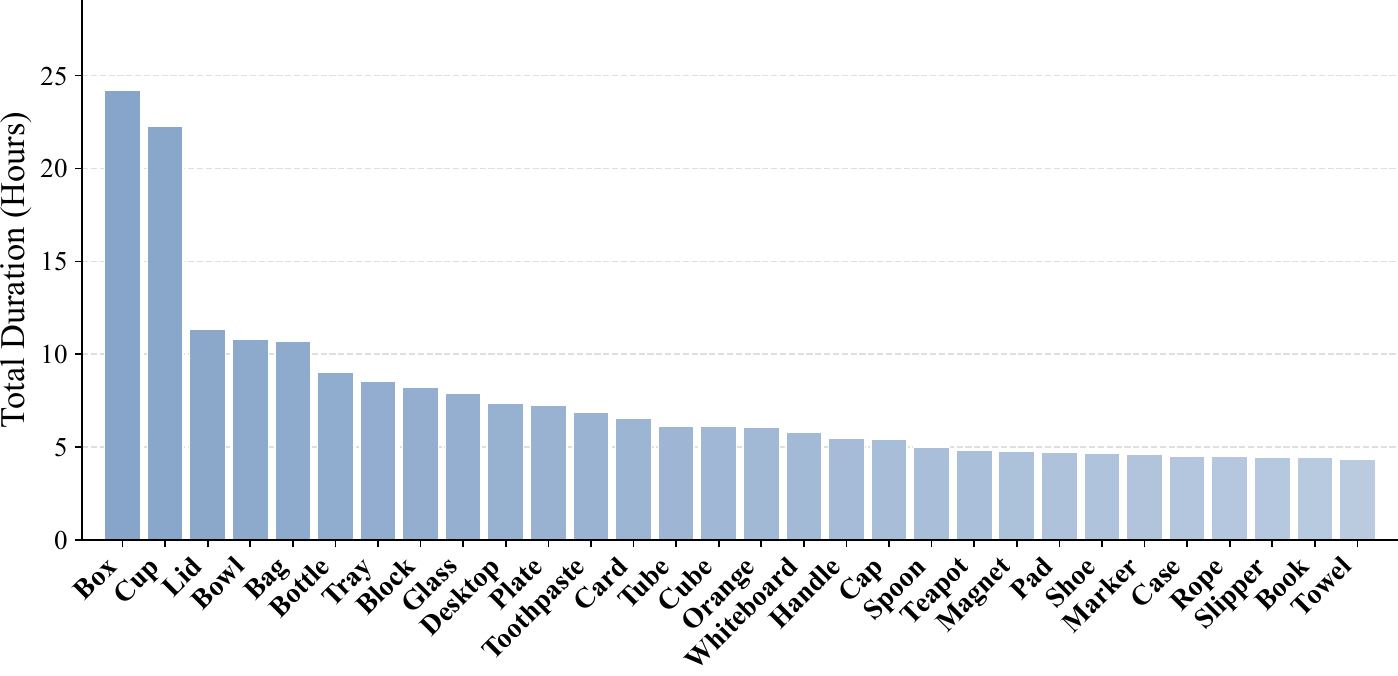}
    \subcaption{Top 30 noun frequencies}
    \label{fig:top30_nouns}
  \end{minipage}

  \vspace{0.5cm}

  \begin{minipage}[t]{0.44\linewidth}
    \centering
    \includegraphics[width=\linewidth]{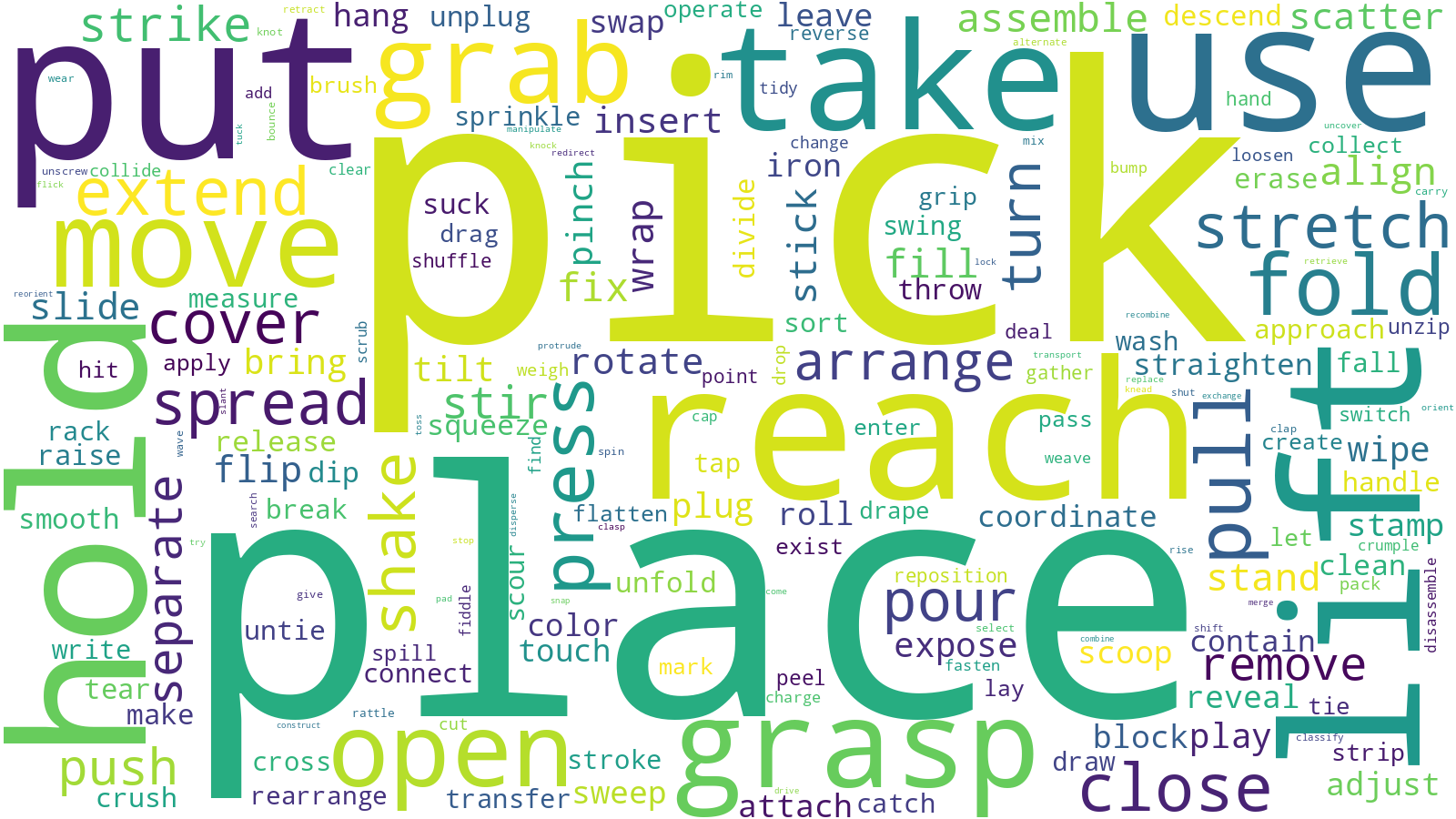}
    \subcaption{Word cloud of verbs in annotation}
    \label{fig:verb}
  \end{minipage}\hfill
  \begin{minipage}[t]{0.52\linewidth}
    \centering
    \includegraphics[width=\linewidth]{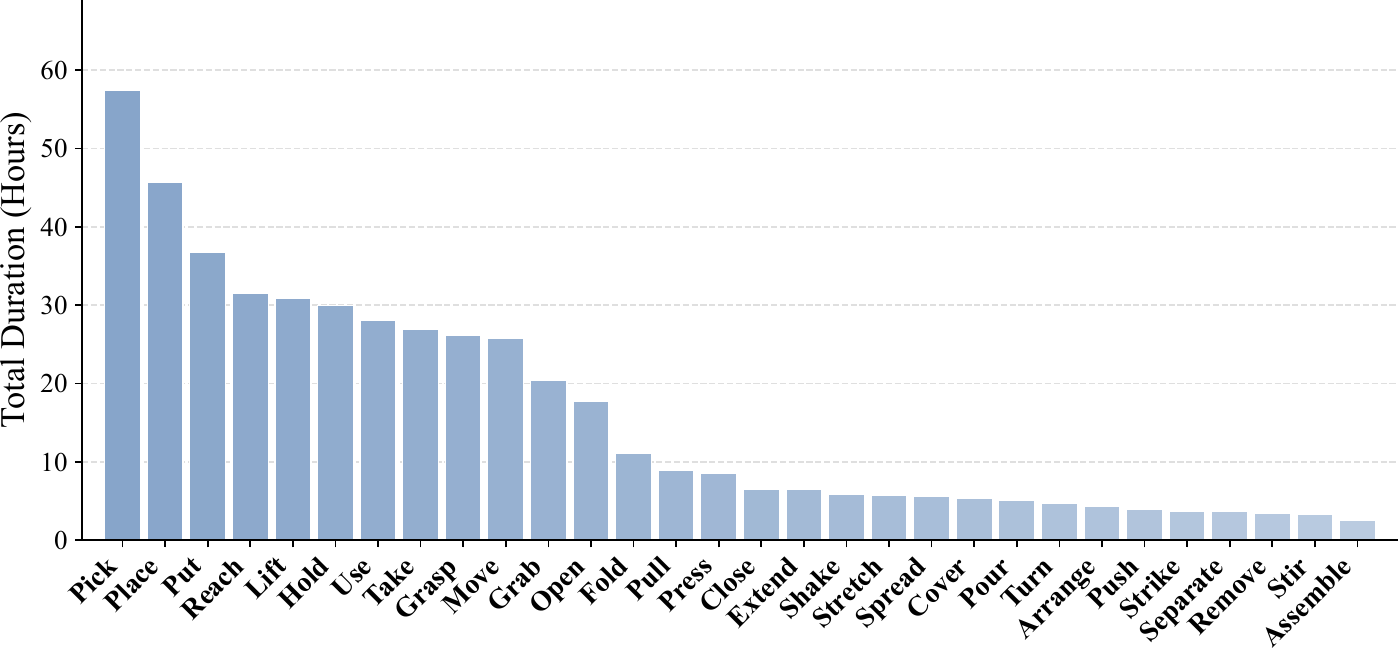}
    \subcaption{Top 30 verb frequencies}
    \label{fig:top30_verbs}
  \end{minipage}

  \vspace{0.5cm}

  \begin{minipage}[t]{0.42\linewidth}
    \centering
    \includegraphics[width=\linewidth]{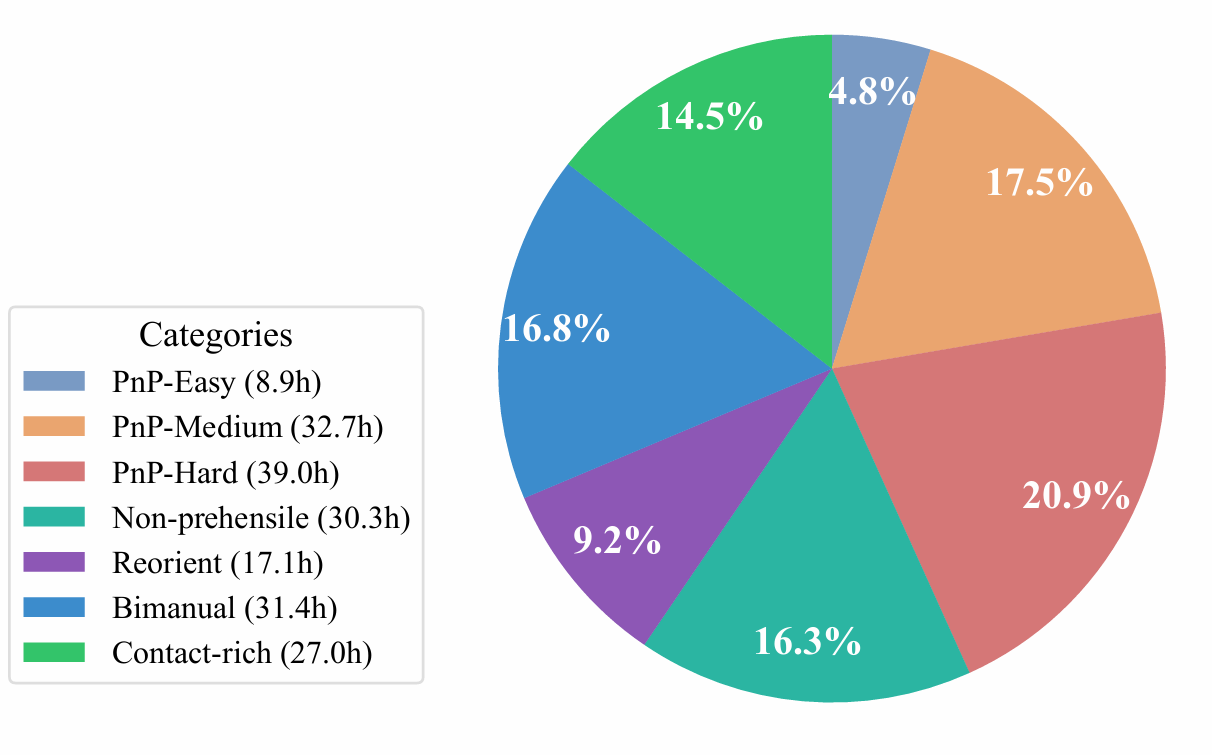}
    \subcaption{Task duration by seven categories}
    \label{fig:duration_pie}
  \end{minipage}\hfill
  \begin{minipage}[t]{0.57\linewidth}
    \centering
    \includegraphics[width=\linewidth]{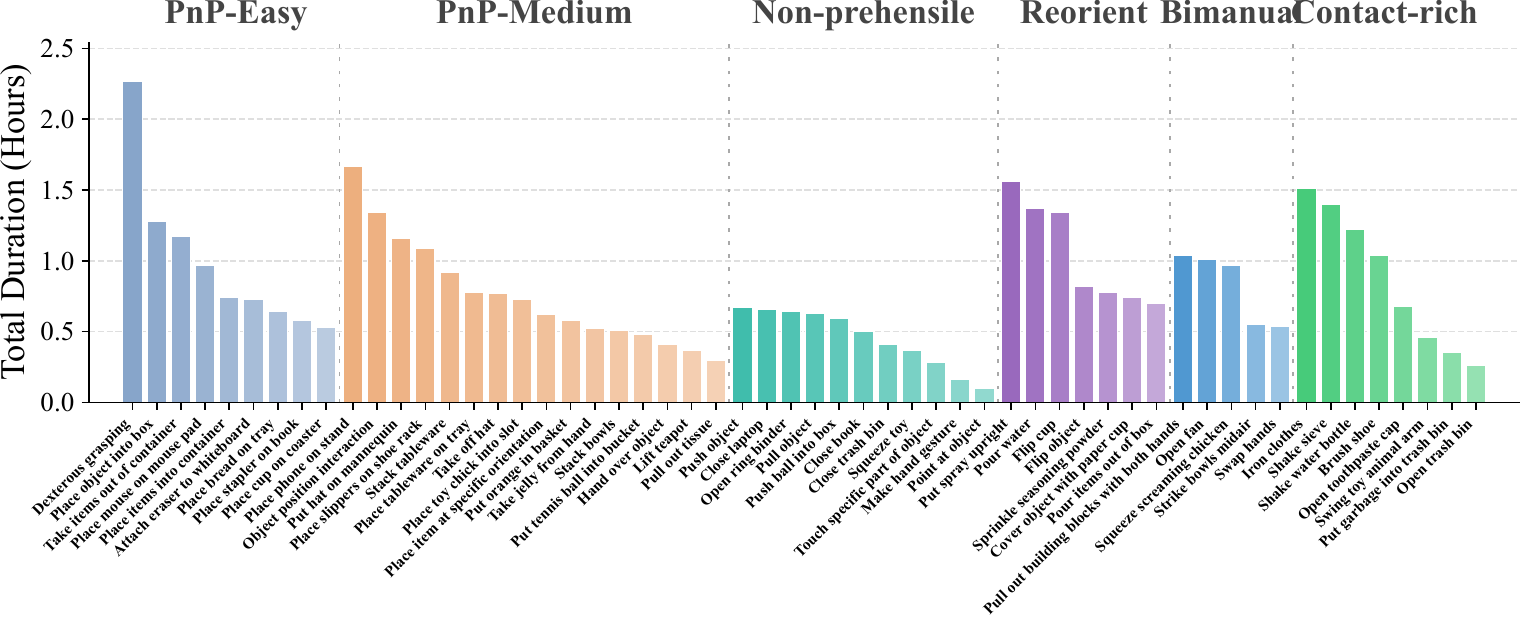}
    \subcaption{Common task duration statistics}
    \label{fig:common_task}
  \end{minipage}

  \vspace{0.5cm}

  \begin{minipage}[t]{1.00\linewidth}
    \centering
    \includegraphics[width=\linewidth]{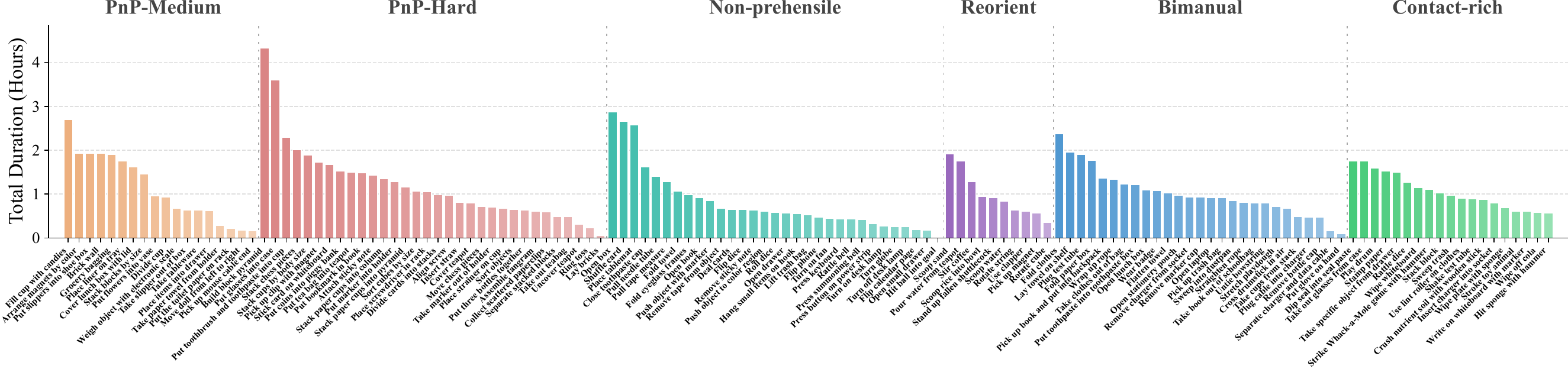}
    \subcaption{Long-tail task duration statistics}
    \label{fig:long_tail_task}
  \end{minipage}

  \caption{Teleoperation dataset statistics. (1) \Cref{fig:noun} - \Cref{fig:top30_verbs}: Word clouds and top-30 frequency distributions of nouns and verbs in language annotations. (2) \Cref{fig:duration_pie}: Task duration breakdown by seven manipulation categories (PnP-Easy/Medium/Hard, non-prehensile, reorient, bimanual, contact-rich). (3) \Cref{fig:common_task} and \Cref{fig:long_tail_task}: Detailed duration statistics for common tasks (56 tasks) and long-tail tasks (137 tasks).}
  \label{fig:teleoperation_all_charts}
\end{figure}

\begin{figure}[t]
  \centering
  \includegraphics[width=0.95\linewidth]{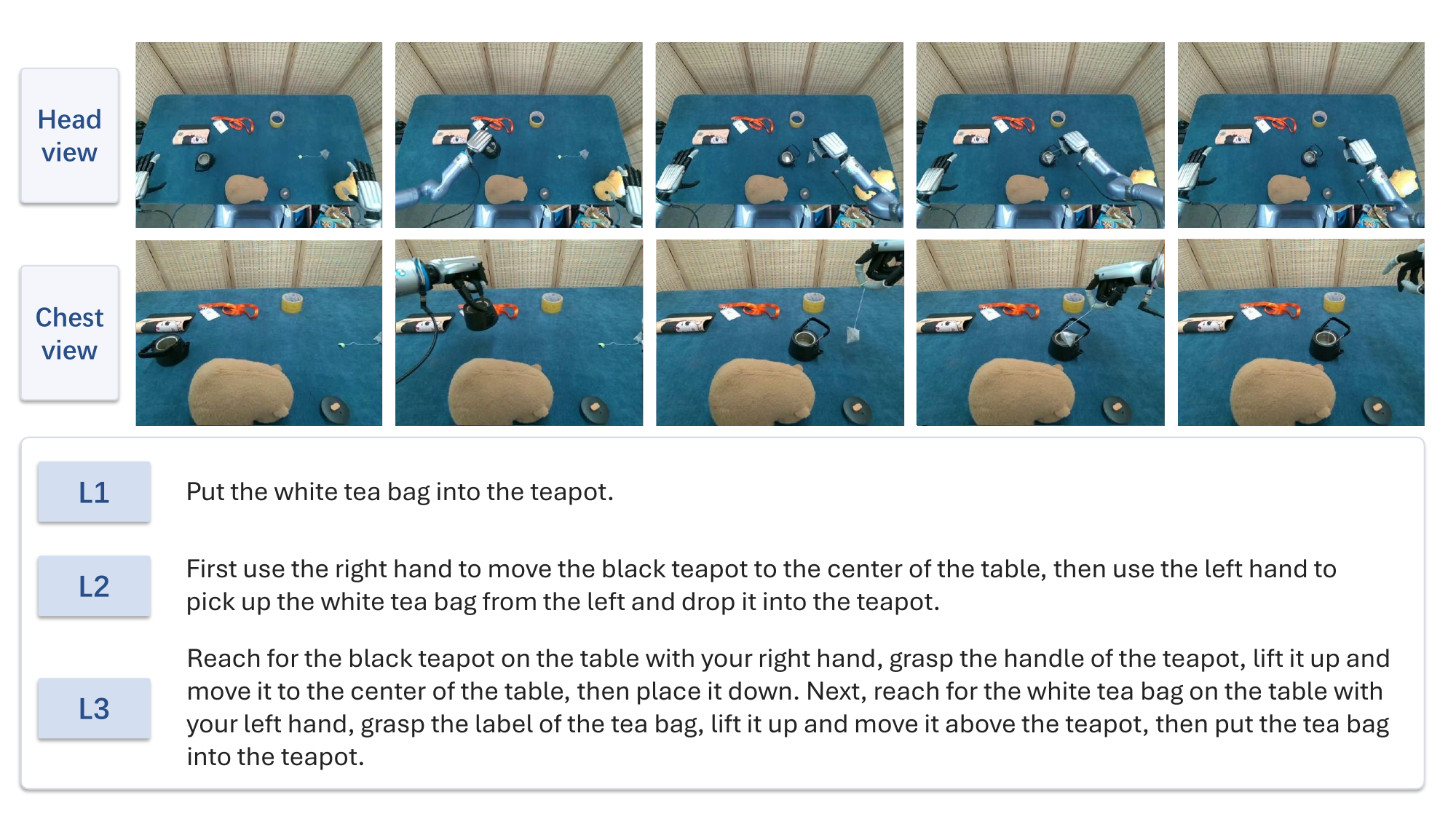}
\caption{\textbf{Dual-view image sequence and three-level language annotations of a representative trajectory}. The frames illustrate a cluttered, randomized setup where the operator executes the task in a natural, human-like manner. The hierarchical language annotations describe the manipulation process from coarse to fine, assisting the model in aligning free-form instructions with physical actions.}
  \label{fig:teleop-episode}
\end{figure}

\begin{figure}[t]
  \centering
  \includegraphics[width=0.95\linewidth]{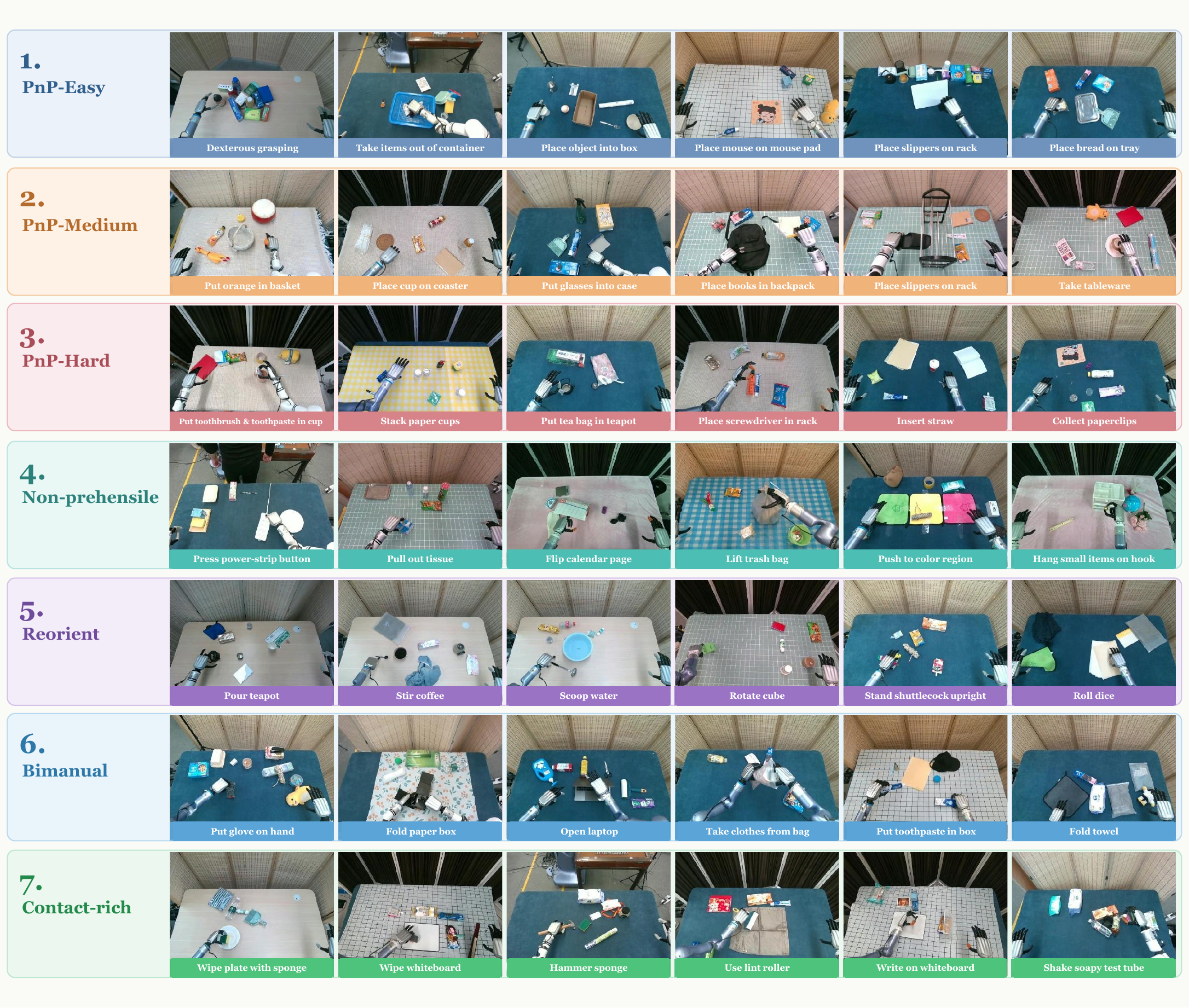}
\caption{\textbf{Representative task examples across seven manipulation categories}. These diverse tasks are executed in cluttered, randomized scenarios using natural, human-like manipulations, covering a wide range of action semantics and manipulation primitives to facilitate the efficient grounding of pre-trained egocentric human priors onto the physical robot.}
\label{fig:teleop-tasks}
\end{figure}

As illustrated in \Cref{fig:teleoperation_all_charts}, we analyze the $187$-hour real-robot dataset across several key dimensions, including the verb and noun word clouds with their top 30 frequency distributions, the task duration breakdown across seven manipulation categories, and the detailed duration statistics for the $56$ common and $137$ long-tail tasks. These statistics reveal a diverse vocabulary of actions and manipulated objects, alongside a balanced distribution across both manipulation categories and task durations. Encompassing rich semantic concepts and manipulation primitives, this dataset effectively grounds pre-trained human manipulation priors onto the RealMan embodiment. Additionally, \Cref{fig:teleop-episode} showcases the dual-view image sequences and language annotations of a teleoperation trajectory, while \Cref{fig:teleop-tasks} visualizes sample tasks across each category.

\clearpage

\begin{table}[t]
\caption{\textbf{VLM datasets used for EgoSteer co-training}. This table details the sample sizes and the multi-modal and embodied knowledge domains covered by each dataset. This co-training mixture integrates general vision-language knowledge with interaction understanding and spatial-geometric priors, preserving the model's inherent world knowledge while facilitating its ability to follow free-form instructions to generalize across diverse manipulation tasks.}
  \centering
  \renewcommand{\arraystretch}{1.3}
  \resizebox{\linewidth}{!}{%
    \begin{tabular}{lcccccccccc}
    \toprule
    \textbf{Dataset}
    & \makecell{\textbf{Samples}}
    & \makecell{\textbf{Captioning}}
    & \makecell{\textbf{Visual Question}\\\textbf{Answering}}
    & \makecell{\textbf{Multiple-Choice}\\\textbf{Question}}
    & \makecell{\textbf{Pointing}}
    & \makecell{\textbf{Bounding Box}}
    & \makecell{\textbf{Affordance}}
    & \makecell{\textbf{Trajectory}}
    & \makecell{\textbf{Spatial}}
    & \makecell{\textbf{Planning}} \\
    \midrule
    \rowcolor{zebra} \textbf{FineVision}~\citep{wiedmann2025finevision} & $3.5\text{M}$     & \cmark & \cmark & \cmark &        & \cmark &        &        &        &        \\
    \textbf{RefSpatial}~\citep{zhou2026roborefer}  & $2.5\text{M}$  &        & \cmark & \cmark & \cmark &        &        &        & \cmark &        \\
    \rowcolor{zebra} \textbf{RoboInter-VQA}~\citep{li2026robointer} &  $1.6\text{M}$   &   & \cmark & \cmark & \cmark & \cmark &        & \cmark & \cmark & \cmark \\
    \textbf{RoboPoint}~\citep{yuan2024robopoint}     &   $1.3\text{M}$   &  & \cmark &        & \cmark & \cmark &        &        & \cmark &        \\
    \rowcolor{zebra} \textbf{RoboAfford}~\citep{tang2025roboafford} & $765\text{K}$   &        & \cmark &        & \cmark & \cmark & \cmark &        &        &        \\
    \textbf{Robo2VLM}~\citep{chen2025robo2vlm}   & $678\text{K}$   &        & \cmark & \cmark &        &        &        &        &        &        \\
    \rowcolor{zebra} \textbf{ShareRobot}~\citep{ji2025robobrain}    &   $13\text{K}$   &  & \cmark &        &        & \cmark & \cmark & \cmark &        & \cmark \\
    \bottomrule
    \end{tabular}%
  }
  \vspace{8pt}

  \vspace{-15pt}
  \label{tab:vlm-dataset-info}
\end{table}

\section{Details of EgoSteer}
\label[appendix]{app:egosteer}

\begin{wrapfigure}{r}{0.4\textwidth}
  \centering
  \vspace{-20pt} 
  \includegraphics[width=\linewidth]{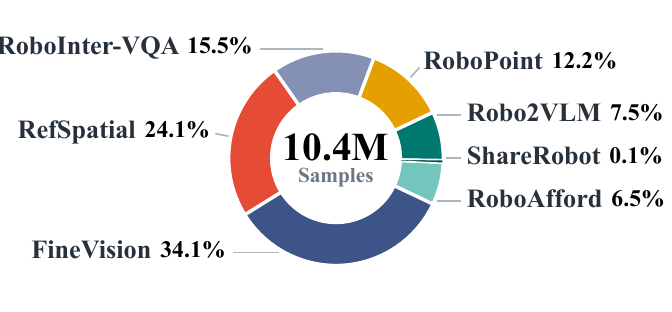}
  \caption{Statistical distribution of samples across VLM co-training datasets.}
  \label{fig:vlm-dataset-composition}
  \vspace{-20pt}
\end{wrapfigure}

This section first introduces the VLM datasets utilized for co-training with VLA data, \emph{i.e.} egocentric human videos and real-robot data, to preserve EgoSteer's vision-language knowledge and ensure generalization (\Cref{app:vlm-cotraining}). We then present implementation details of EgoSteer (\Cref{app:egosteer-details}).

\subsection{VLM Co-Training Data}
\label[appendix]{app:vlm-cotraining}

To preserve general vision-language reasoning capabilities while simultaneously cultivating robust robotic task comprehension, a $10.4\text{M}$-sample VLM co-training mixture is curated across seven datasets, ranging from open-world perception to embodied interaction grounding, to co-train EgoSteer. This mixture comprises four categories of data:
\begin{itemize}[leftmargin=*, noitemsep, topsep=0pt]
    \item \textbf{General VLM Pre-Training}: Incorporates FineVision~\citep{wiedmann2025finevision} to prevent the catastrophic forgetting of open-world semantic concepts and preserve general visual-language reasoning.
    \item \textbf{Spatial Grounding}: Utilizes RefSpatial~\citep{zhou2026roborefer} and RoboPoint~\citep{yuan2024robopoint} for multi-step spatial referring, 2D visual grounding, and precise coordinate localization.
    \item \textbf{Embodied QA}: Integrates RoboInter-VQA~\citep{li2026robointer}, Robo2VLM~\citep{chen2025robo2vlm}, and ShareRobot~\citep{ji2025robobrain} to support embodied question answering and temporal reasoning. This preserves the model's capabilities in high-level task planning and causal scene understanding.
    \item \textbf{Affordance Perception}: Adopts RoboAfford~\citep{tang2025roboafford} for fine-grained manipulation affordance prediction and spatial interaction grounding.
\end{itemize}

The sample distribution of these datasets is illustrated in \Cref{fig:vlm-dataset-composition}, with their covered multimodal, embodied knowledge domains detailed in \Cref{tab:vlm-dataset-info}. We standardize these datasets to comply with the conversational input format of Qwen3-VL. Specifically, we normalize 2D bounding box and point coordinates to Qwen3-VL's native $[0, 1000]$ scale and adopt its standard representation formats. To ensure training stability, we only utilize single-image samples and exclude those with context lengths exceeding our maximum budget. By co-training on this VLM mixture, EgoSteer preserves open-world vision-language understanding and reasoning while enhancing its comprehension of robotics-specific tasks, thereby assisting the policy in following open-ended instructions and generalizing to novel manipulation scenarios.

\subsection{Implementation Details}
\label[appendix]{app:egosteer-details}

\paragraph{Backbone Input Scheme.} EgoSteer treats the image observation history as a temporal video sequence to leverage the native video-processing capabilities of the Qwen3-VL-2B backbone. In practice, this history is downsampled to $6$ frames at $1$\,FPS, covering a $5$\,s window, with proprioceptive states sampled at the corresponding timestamps. Language instructions and camera intrinsics are formatted as textual inputs, while the proprioceptive state history is encoded by a two-layer MLP and injected as continuous tokens. Because this proprioceptive history correlates strongly with target actions, the model is susceptible to shortcut learning, tending to ignore visual inputs and task instructions in favor of proprioception. To mitigate this, each frame of the proprioceptive history is replaced by a learnable mask token with a $75\%$ probability during training, forcing the model to attend to the full multimodal context. Additionally, when utilizing dual-camera inputs during real-robot post-training, the chest-camera sequence is randomly dropped with a $50\%$ probability to prevent over-reliance on chest-view observations.

\paragraph{Action Expert.} The action expert is built on a DiT architecture, comprising $14$ layers with a hidden dimension of $1024$ and an intermediate size of $2816$. It features $8$ attention heads, each with a head dimension $d_{\text{head}} = 128$, totaling approximately $300$\,M parameters. 

The action expert operates on continuous action chunks of length $h = 32$ at a frequency of $30$~Hz. Consistent with the VLM backbone, we apply Interleaved MRoPE for positional encoding. During pre-training, we set the delay $d = 0$ to disable training-time RTC, maximizing action supervision signals and learning rich human manipulation priors. During real-robot post-training, the simulated delay $d$ is uniformly sampled as $d \sim \mathcal{U}([0, 5])$ to accommodate varying inference latencies during deployment. For the flow matching timestep $\eta \in [0, 1]$, the prefix action $\mathbf{a}_{\text{pre}}$ is assigned $\eta = 1$, representing no noise and excluding it from the loss computation. Conversely, the target suffix action $\mathbf{a}_{\text{suf}}$ has its timestep sampled from the probability distribution $P(\eta) = \text{Beta}(\frac{s - \eta}{s}; 1.5, 1)$ with $s = 0.999$~\citep{black2026pi0visionlanguageactionflowmodel}. This timestep $\eta$ is sinusoidally encoded, mapped through a two-layer MLP, and injected into the action expert via AdaLN-Zero.

Layer $\ell$ of the action expert employs a joint attention mechanism, attending to both itself and the VLM backbone's key-value cache from layer $f(\ell) = 2\ell$ after a linear projection. Specifically, let the query, key, and value of the $m$-th attention head in layer $\ell$ of the action expert be $\mathbf{Q}_{\ell, m}^{\text{AE}}, \mathbf{K}_{\ell, m}^{\text{AE}}, \mathbf{V}_{\ell, m}^{\text{AE}} \in \mathbb{R}^{h \times d_{\text{head}}}$, and the corresponding key and value of the $m$-th head in layer $f(\ell)$ of the backbone be $\mathbf{K}_{f(\ell), m}^{\text{B}}, \mathbf{V}_{f(\ell), m}^{\text{B}} \in \mathbb{R}^{N_{\text{B}} \times d_{\text{head}}}$, where $N_{\text{B}}$ denotes the backbone's input sequence length. The joint attention at layer $\ell$ for the $m$-th head is mathematically formulated as:
\begin{equation}
\text{Softmax}\left(\frac{1}{\sqrt{d_{\text{head}}}}\mathbf{Q}_{\ell, m}^{\text{AE}} \left(\text{concat}[\mathbf{K}_{f(\ell), m}^{\text{B}}\mathbf{W}_{\ell}^{\text{K}}, \mathbf{K}_{\ell, m}^{\text{AE}}]\right)^{\text{T}}\right)\text{concat}[\mathbf{V}_{f(\ell), m}^{\text{B}}\mathbf{W}_{\ell}^{\text{V}}, \mathbf{V}_{\ell, m}^{\text{AE}}],
\label{eq:joint_attention}
\end{equation}
where $\mathbf{W}_\ell^{\text{K}}, \mathbf{W}_\ell^{\text{V}} \in \mathbb{R}^{d_{\text{head}} \times d_{\text{head}}}$ represent learnable projection matrices. This linear projection on the backbone's key and value representations is designed to align the semantic spaces of the backbone and the action expert.

During real-world deployment, the simulated delay is set to $d=4$ to cover physical inference latency. To achieve efficient closed-loop control, only the first $12$ steps of the predicted $32$-step action chunk are retained. Subtracting the $4$ prefix steps used for latency conditioning, the robot actually executes $8$ new action steps per inference cycle. This high-frequency, asynchronous execution enables highly responsive control, rendering the system robust to dynamic manipulation tasks.

\paragraph{World Model Expert.} The world model expert is a lightweight Transformer comprising $4$ layers, with its single-layer architecture identical to Qwen3's text layer. It has a hidden dimension of $1024$, an intermediate size of $4096$, and $8$ attention heads with a head dimension $d_{\text{head}} = 128$, totaling approximately $70$\,M parameters. At layer $\ell$, the world-model expert employs a joint attention mechanism, attending to both itself and the VLM backbone's key-value cache from layer $f(\ell) = 7\ell$ after a linear projection, consistent with the formulation of the action expert.

For inputs, the relative camera motion $\Delta\mathbf{T} \in SE(3)$ is flattened into a $16$-dimensional vector and encoded into a single continuous token via a two-layer MLP. Let the ground-truth DINOv3 features of the future frame $\mathbf{I}_{t+h-1}$ be $\mathbf{Z} \in \mathbb{R}^{H_v \times W_v \times C_{\text{DINO}}}$. We utilize DINOv3 (ViT-L/16)~\citep{simeoni2025dinov3} for feature extraction with an input resolution of $384\times384$, yielding a spatial resolution of $H_v = W_v = 24$ and a feature dimension of $C_{\text{DINO}} = 1024$. To align with the spatial token-merge format of the backbone, the sequence length $L_{\mathbf{z}}$ of the input query vector $\mathbf{z}$ is configured to match the merged spatial resolution:
$L_{\mathbf{z}} = H'_v \times W'_v = \frac{H_v}{2} \times \frac{W_v}{2} = 144 $. The world model expert outputs $\hat{\mathbf{Y}} \in \mathbb{R}^{H'_v \times W'_v \times d_{\text{WM}}}$ with a channel dimension $d_{\text{WM}} = 1024$, which is subsequently mapped back to the original DINOv3 spatial resolution through a $2 \times 2$ linear upsampling projection layer, yielding the reconstructed feature map $\hat{\mathbf{Z}} \in \mathbb{R}^{H_v \times W_v \times C_{\text{DINO}}}$. The world model objective is optimized via a mean squared error (MSE) loss:
\begin{equation}
\mathcal{L}_{\text{WM}} = \frac{1}{H_v \cdot W_v \cdot C_{\text{DINO}}} \sum_{u=1}^{H_v} \sum_{v=1}^{W_v} \left\| \mathbf{Z}_{u, v} - \hat{\mathbf{Z}}_{u, v} \right\|_2^2.
\label{eq:wm_loss}
\end{equation}

Compared to direct image prediction in pixel space, regressing DINOv3 features preserves rich semantic information, naturally filtering out lighting variations and background noise to provide more stable gradient guidance for the VLM backbone. Furthermore, the world model expert adopts Interleaved MRoPE positional encoding, consistent with the VLM backbone, which enhances spatial-temporal awareness of multimodal sequences.

\paragraph{Attention Pattern.} The Qwen3-VL backbone employs causal attention. Both the action expert and the world-model expert jointly attend to their entire respective sequences and the entire backbone sequence. Crucially, the action expert and the world-model expert do not attend to each other.

\paragraph{Data Processing.} Due to significant variations in scale and quality across the 12 egocentric pre-training datasets, a heuristic sampling weight scheme is employed to balance different data sources. Specifically, each dataset $i$ is assigned a subjective quality score $w_i \in [1, 10]$ based on data quality. To mitigate scale discrepancies, this score is scaled by the square root of the total frame count $n_i$, yielding the final sampling weight $W_i = w_i \cdot n_i^{0.5}$. For data augmentation, ColorJitter is applied to the input images. Furthermore, all action dimensions, except for wrist rotations, are normalized to the range $[-1, 1]$ using their 1st and 99th percentiles estimated from a randomly sampled subset.

\paragraph{Joint Optimization Objective.} To balance the primary flow-matching task and the two auxiliary targets, the total training loss $\mathcal{L}_{\text{total}}$ is defined as:
\begin{equation}
\mathcal{L}_{\text{total}} = \mathcal{L}_{\text{CFM}} + \mathcal{L}_{\text{WM}} + 0.05 \mathcal{L}_{\text{VLM}},
\label{eq:total_loss}
\end{equation}
where $\mathcal{L}_{\text{CFM}}$ is the action flow-matching loss, $\mathcal{L}_{\text{WM}}$ is the world-model feature regression loss, and $\mathcal{L}_{\text{VLM}}$ is the autoregressive next-token prediction loss of the VLM. The loss weights are set to align the numerical scales of the three loss terms.

\paragraph{Training Infrastructure.}
Both the pre-training corpus curated by EgoSmith and the real-robot dataset are stored sequentially as individual episodes in the WebDataset format. During data loading, training batches are randomly sampled from a large shuffle buffer of size $16,384$. This buffer is continuously filled by drawing samples from each dataset proportionally to its pre-defined weight. For each dataset, the data stream is maintained by randomly selecting and streaming WebDataset shards, caching samples via a sliding window, and applying a $20\%$ random retention probability. By combining randomized shard reading, sample dropping, and large-buffer shuffling, this scheme ensures training randomness while leveraging the sequential streaming of WebDataset, thereby drastically reducing I/O pressure. In addition to WebDataset, we also support the LeRobot v3 format using the same pipeline, randomly selecting and sequentially streaming episodes instead of shards. Furthermore, in multi-node distributed training, we explicitly manage Python's garbage collection to effectively mitigate training speed fluctuations and synchronization jitter across the cluster, ensuring high efficiency and stability.

\clearpage
\section{Experimental Details}
\label[appendix]{app:experiments}

\begin{figure*}[!t]
    \centering
    \begin{subfigure}[t]{\textwidth}
        \includegraphics[width=\textwidth,page=1]{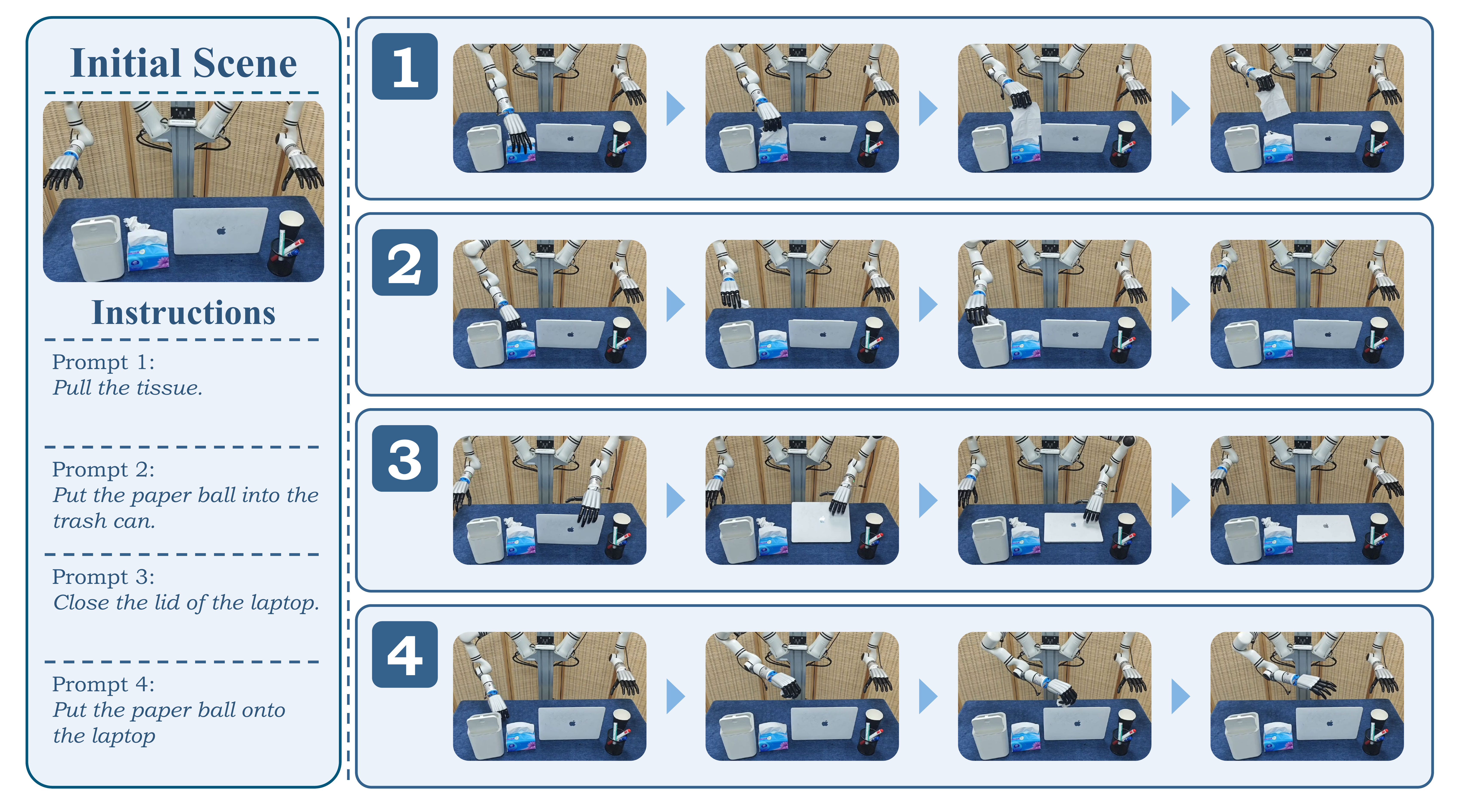}
        \captionsetup{skip=3pt}
        \caption{Steerability in an office scene.}
        \label{fig:steerability-g1}
    \end{subfigure}
    \begin{subfigure}[t]{\textwidth}
        \includegraphics[width=\textwidth,page=2]{figures/steerability_illustration.pdf}
        \captionsetup{skip=3pt}
        \caption{Steerability among objects with similar colors in the same scene.}
        \label{fig:steerability-g2}
    \end{subfigure}
    \caption{Illustrations of the steerability of EgoSteer. It is capable of executing different actions given different language instructions under the same initial scene configuration.}
    \label{fig:steerability}
\end{figure*}

\begin{figure*}[h]
    \centering
    \begin{subfigure}[t]{\textwidth}
        \includegraphics[width=\textwidth,page=1]{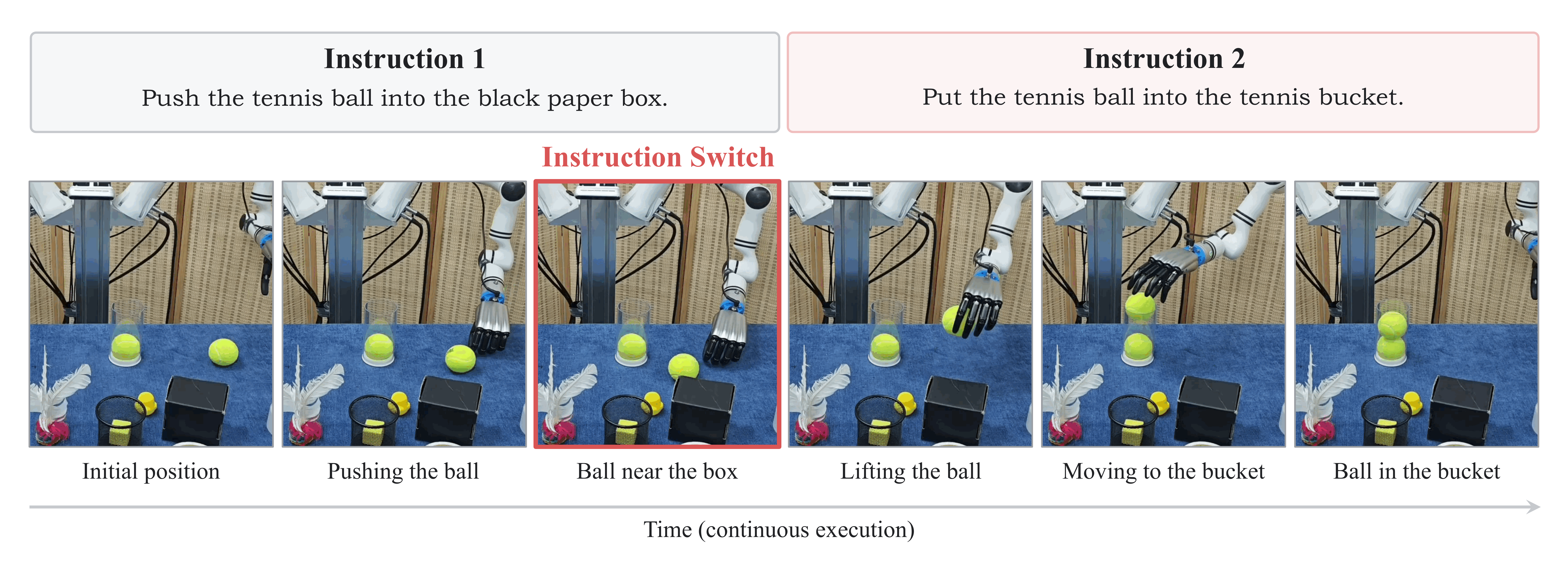}
        \captionsetup{skip=3pt}
        \caption{The robot initially pushes the tennis ball toward the box. After the instruction switch, it redirects the ball to the bucket.}
        \label{fig:resteer-g1}
    \end{subfigure}
    \begin{subfigure}[t]{\textwidth}
        \includegraphics[width=\textwidth,page=2]{figures/resteer_illustration.pdf}
        \captionsetup{skip=3pt}
        \caption{The robot initially moves the yellow cube toward the recycle bin. After the instruction switch, it redirects the cube to the metal plate.}
        \label{fig:resteer-g2}
    \end{subfigure}
    \caption{Illustrations of EgoSteer handling instruction switching during continuous execution.
    The instruction is switched at an intermediate timestep, after which EgoSteer adjusts its subsequent behavior to follow the updated instruction. This indicates that EgoSteer is genuinely following language guidance rather than memorizing spurious scene-task correlation. The red box indicates the switching moment.}
    \label{fig:resteer}
\end{figure*}

\begin{figure}[ht]
    \centering
    \begin{minipage}{0.80\linewidth}
        \centering
        \newsavebox{\realmanSetupPanel}
        \sbox{\realmanSetupPanel}{%
            \begin{minipage}{0.485\linewidth}
                \centering
                \begin{minipage}[t]{0.485\linewidth}
                    \centering
                    \vspace{0pt}
                    {\footnotesize Box Folding\par}
                    \vspace{2pt}
                    \includegraphics[width=\linewidth]{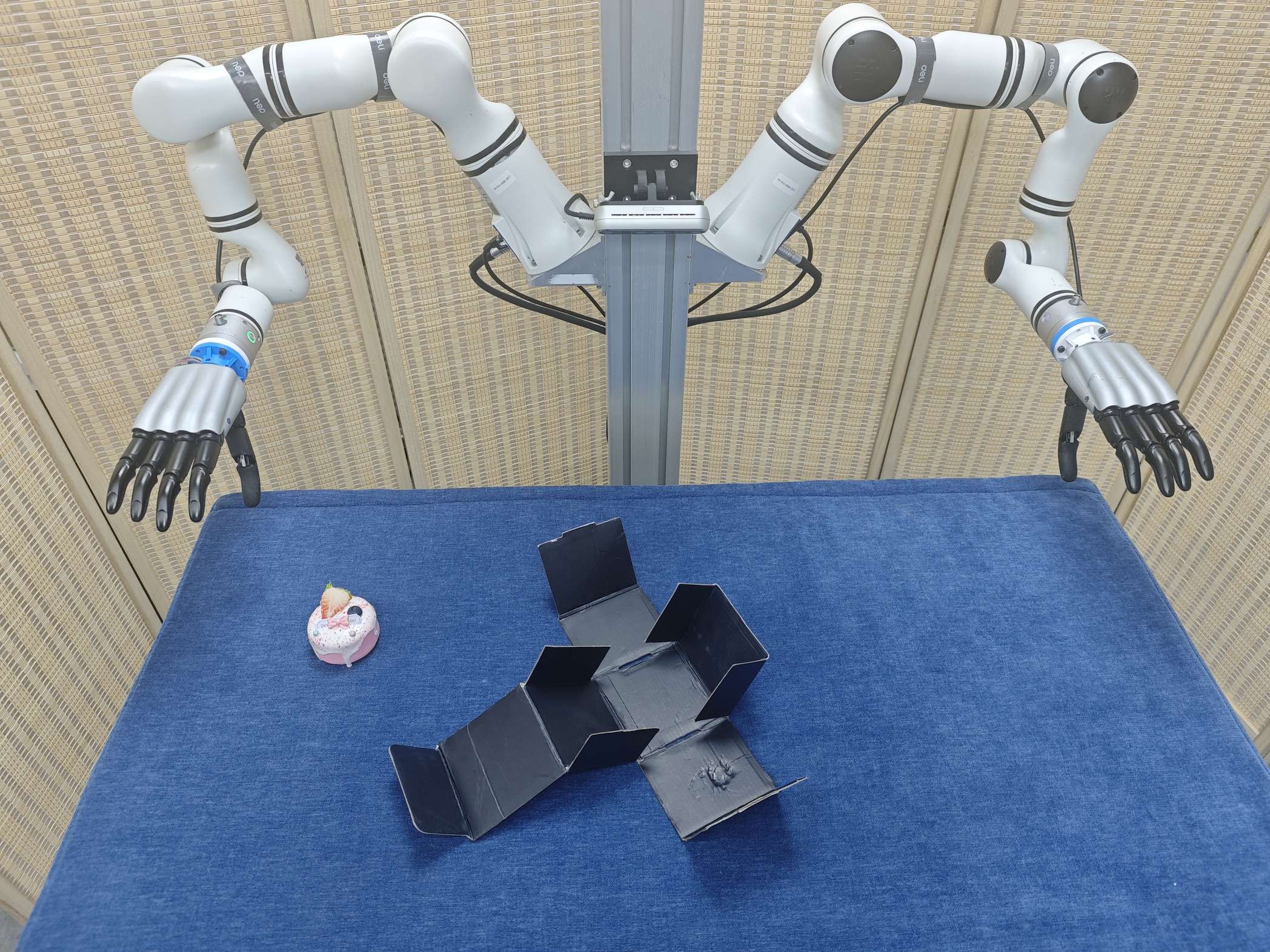}
                \end{minipage}%
                \hfill
                \begin{minipage}[t]{0.485\linewidth}
                    \centering
                    \vspace{0pt}
                    {\footnotesize Sport Stacking\par}
                    \vspace{2pt}
                    \includegraphics[width=\linewidth]{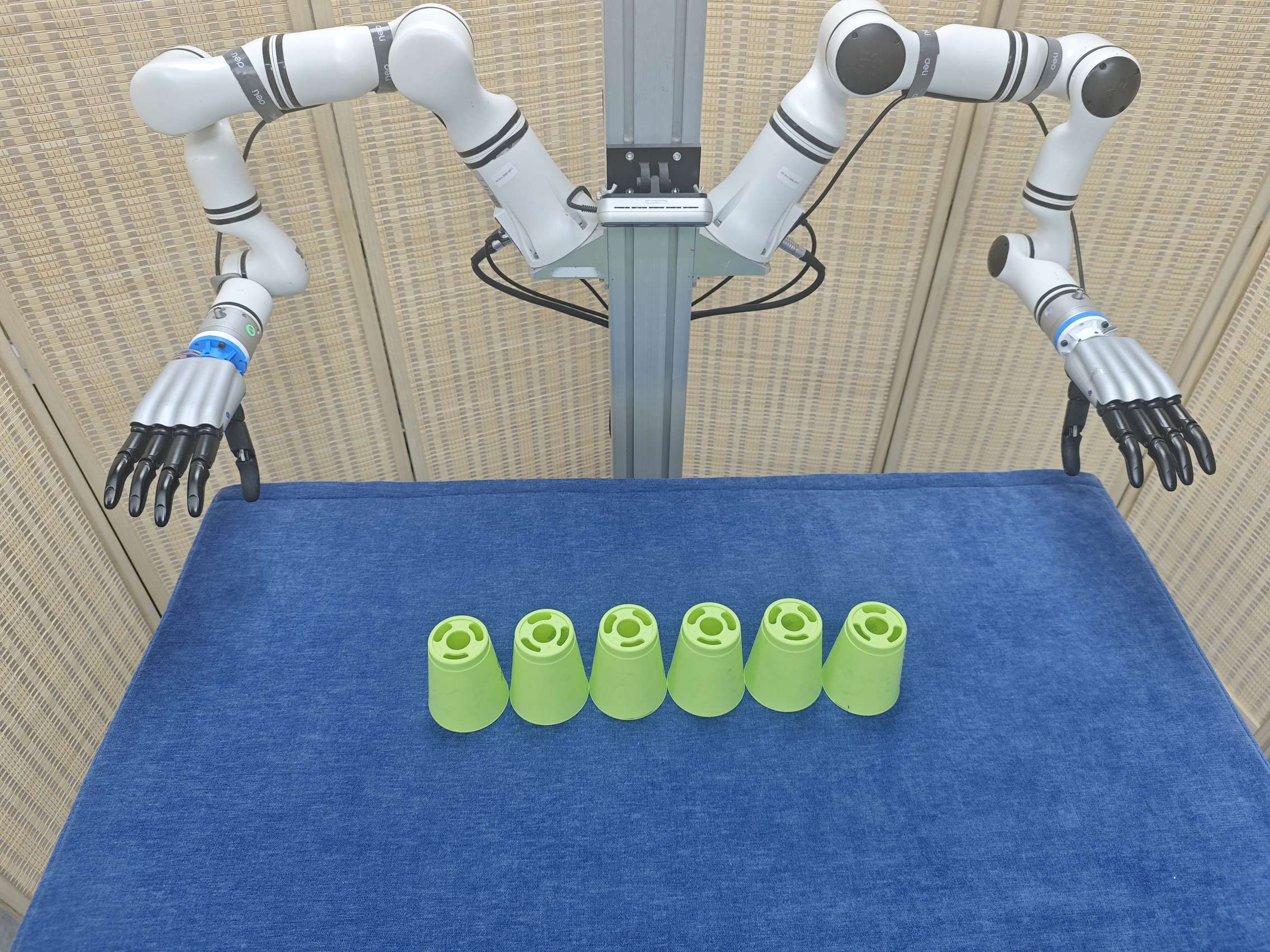}
                \end{minipage}
    
                \vspace{7pt}
    
                \begin{minipage}[t]{0.485\linewidth}
                    \centering
                    \vspace{0pt}
                    {\footnotesize Phone Packing\par}
                    \vspace{2pt}
                    \includegraphics[width=\linewidth]{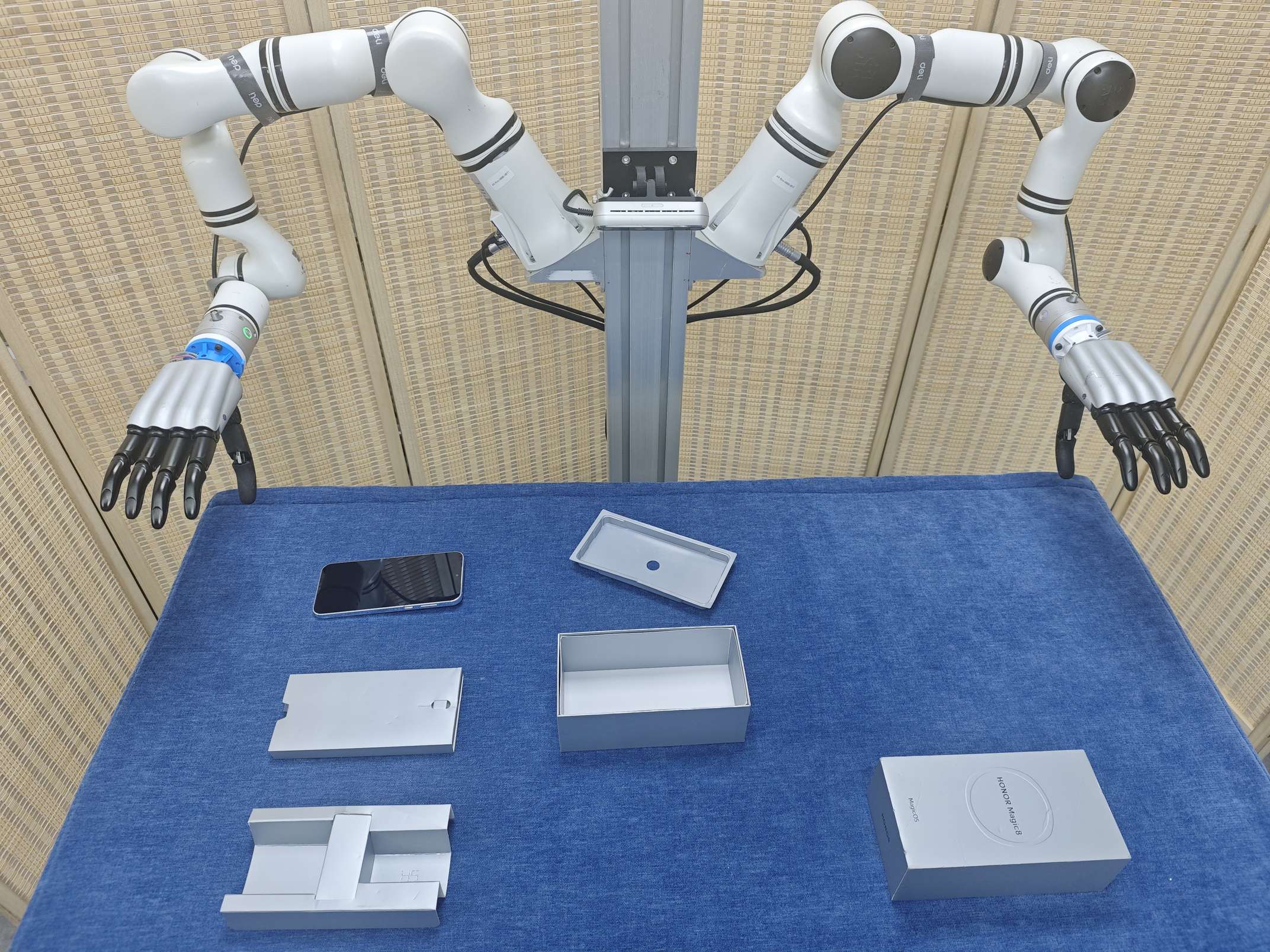}
                \end{minipage}%
                \hfill
                \begin{minipage}[t]{0.485\linewidth}
                    \centering
                    \vspace{0pt}
                    {\footnotesize Ice Cola Making\par}
                    \vspace{2pt}
                    \includegraphics[width=\linewidth]{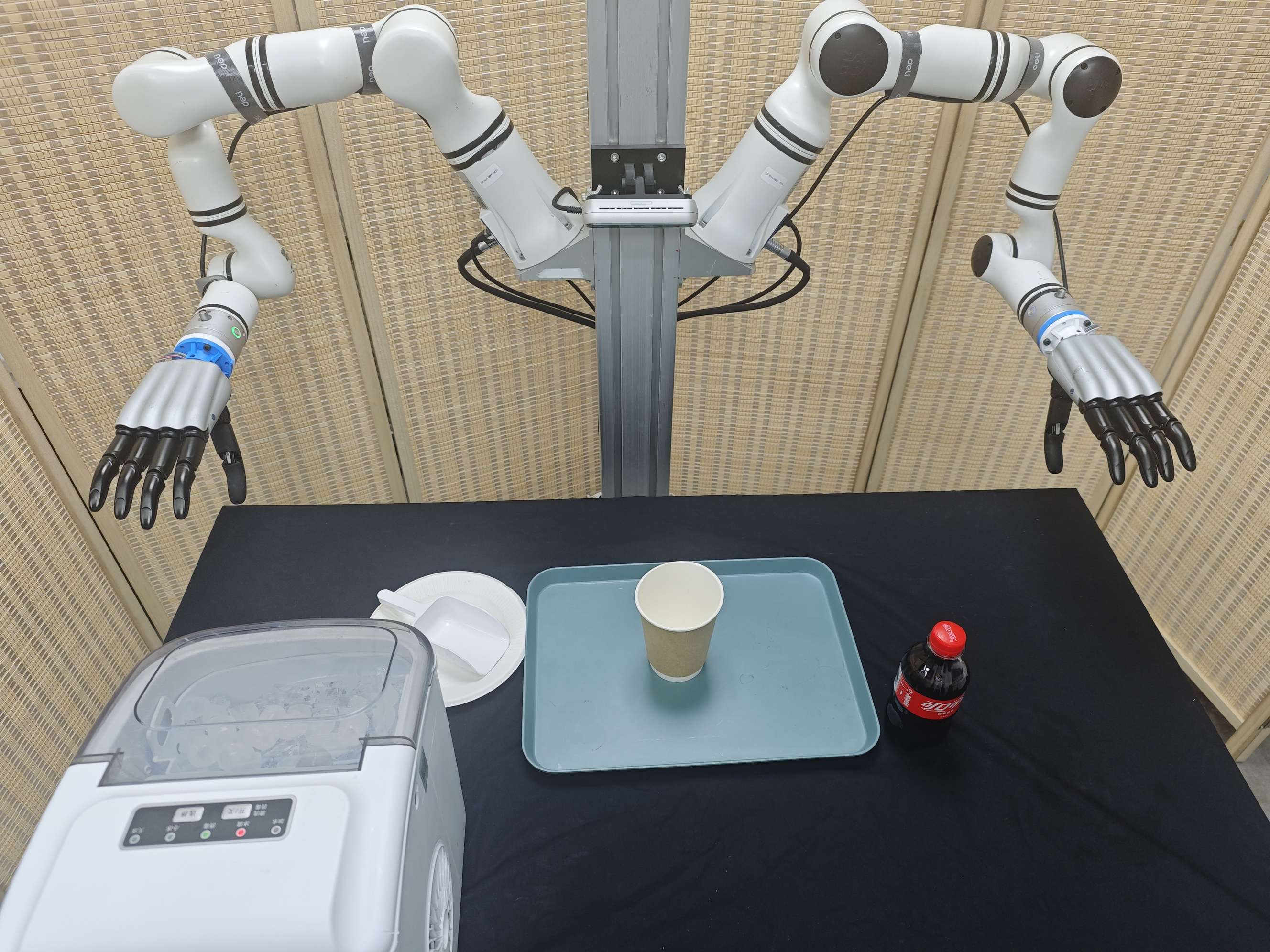}
                \end{minipage}
            \end{minipage}%
        }
        \begin{subfigure}[t]{0.485\linewidth}
            \centering
            \begin{minipage}[c][\dimexpr\ht\realmanSetupPanel+\dp\realmanSetupPanel\relax][c]{\linewidth}
                \centering
                {\footnotesize Cake Unboxing\par}
                \vspace{2pt}
                \includegraphics[width=\linewidth]{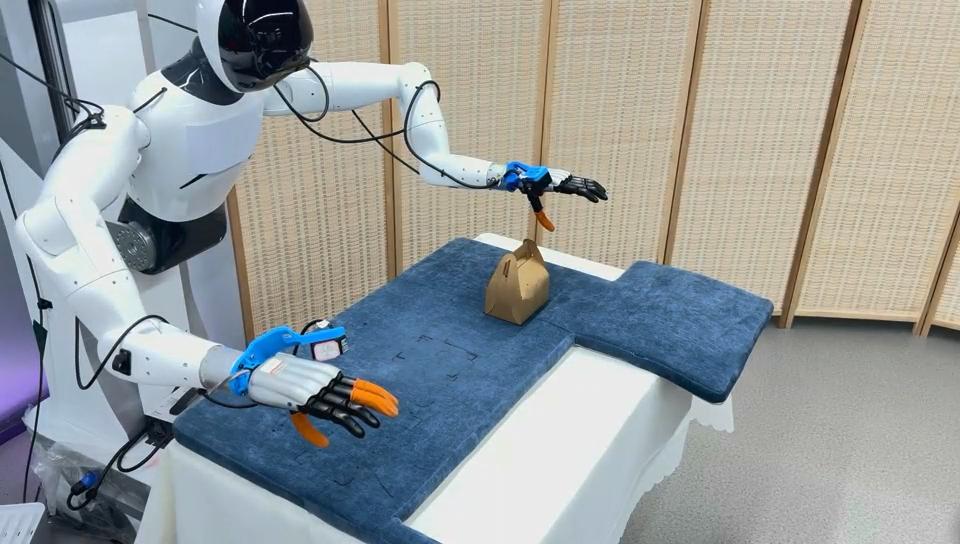}
            \end{minipage}
            \caption{AgiBot G1.}
            \label{fig:single-task-g1}
        \end{subfigure}%
        \hfill
        \begin{subfigure}[t]{0.485\linewidth}
            \centering
            \usebox{\realmanSetupPanel}
            \caption{RealMan.}
            \label{fig:single-task-realman}
        \end{subfigure}
    \end{minipage}
    \caption{Experimental setups for the five single-task evaluations.
    The photographs show the robot embodiments and representative
    tabletop object arrangements.}
    \label{fig:single-task-settings}
\end{figure}

This section provides supplementary details for the experimental evaluations presented in \Cref{sec:experiments}, including the model training hyperparameters, per-task success rates, and additional illustrations. In particular, \Cref{tab:training-config} lists the hyperparameter configurations for both the pre-training and post-training phases in the main experiments of EgoSteer in \Cref{subsec:steerable}. \Cref{tab:dagger_evaluation} presents the per-task success rates on the four evaluation tasks in \Cref{subsec:dagger} to demonstrate the performance gains from DAgger refinement. For the scaling analysis in \Cref{subsec:scaling-and-baseline}, \Cref{tab:scaling-training-config} lists the training hyperparameters of EgoSteer-$3$\,/\,$6$\,/\,$9.6$K-FT models and the baseline trained from scratch, with their corresponding task-specific success rates reported in \Cref{tab:per-task-sr}. \Cref{tab:baseline-post-training} summarizes the post-training configurations of $\pi_{0.5}$ and Being-H0.5  used in our baseline comparisons in \Cref{subsec:scaling-and-baseline}. Additionally, \Cref{tab:cmp-baselines} compares the detailed success rates of our method against $\pi_{0.5}$ and Being-H0.5 across ten tasks. \Cref{tab:ablation-training-config} specifies the training configurations of the ablation experiment from \Cref{subsec:ablation}, including the EgoSteer-$1$K-FT model and its three ablated variants, \textit{No WM-objective}, \textit{No training-RTC}, and \textit{Noisy data}. Their respective performance comparison is provided in \Cref{tab:cmp-ablation}. \Cref{tab:dexterous-single-task} specifies the hyperparameters for EgoSteer-9.6K-FS on the five challenging, long-horizon tasks evaluated in \Cref{subsec:single-task}. Finally, illustrations of EgoSteer's language-following and mid-execution instruction-switching behaviors and few-shot experiment setups are shown in \Cref{fig:steerability,fig:resteer,fig:single-task-settings}.

\begin{table}[!htbp]
\caption{Training configurations for both the pre-training and post-training phases in the main experiments of EgoSteer in \Cref{subsec:steerable}.  
During pre-training, the VLM backbone is frozen for the first 5,000 steps, during which AE and WM are warmed up for 2,000 steps; once it is unfrozen, VLM is warmed up for 2,000 steps.}
\label{tab:training-config}
\centering
\small
\setlength{\tabcolsep}{8pt}
\begin{tabular}{lcc}
\toprule
Hyperparameter & Pre-Training & Post-Training \\
\midrule
Camera setup               & Head             & Head \& Chest               \\
Resolution                 & $384\times384$     & $640\times480$     \\
GPUs                       & 128 A800 & 96 A800 \\
Gradient accumulation      & 2                  & 1                  \\
Global batch size          & 4608               & 384                \\
Training steps             & 175K               & 60K                \\
Learning rate (VLM / AE / WM) & $1{\times}10^{-4}$ / $3{\times}10^{-4}$ / $3{\times}10^{-4}$
                                              & $1{\times}10^{-5}$ / $3{\times}10^{-5}$ / $3{\times}10^{-5}$ \\
Freeze-VLM steps           & 5000                 & 0                 \\
Warmup steps               & 2000                 & 2000                 \\
Training time              & 164\,h             & 29\,h              \\
\bottomrule
\end{tabular}
\vspace{8pt}

\end{table}

\begin{table}[!htbp]
\caption{Success rates across four highly dexterous and failure-prone manipulation tasks, comparing EgoSteer-$9.6$K-DG against EgoSteer-$9.6$K-FT in \Cref{subsec:dagger}. Each task is evaluated over 10 randomized trials. Bold values highlight the best performance.}
\label{tab:dagger_evaluation}
\centering
\small
\setlength{\tabcolsep}{6pt}
\begin{tabular}{lcc}
\toprule
Task & EgoSteer-$9.6$K-DG & EgoSteer-$9.6$K-FT \\
\midrule
Stack tableware      & \textbf{80.0\%} & 50.0\% \\
Close laptop         & \textbf{70.0\%} & 10.0\% \\
Place phone on stand & \textbf{50.0\%} & 0.0\%  \\
Flip cup             & \textbf{50.0\%} & 30.0\% \\
\midrule
\textbf{Average} & \textbf{62.5\%} & 22.5\% \\
\bottomrule
\end{tabular}
\vspace{8pt}

\end{table}

\begin{table}[!htbp]
\caption{Training configurations for the pre-training scaling study in \Cref{subsec:scaling-and-baseline}.}
\label{tab:scaling-training-config}
\centering
\small
\setlength{\tabcolsep}{6pt}
\renewcommand{\arraystretch}{1.15}
\resizebox{\linewidth}{!}{%
\begin{tabular}{l ccc @{\hskip 16pt} cccc}
\toprule
 & \multicolumn{3}{c}{Pre-Training} & \multicolumn{4}{c}{Post-Training} \\
\cmidrule(lr){2-4} \cmidrule(lr){5-8}
Hyperparameter & EgoSteer-$3$K-PT & EgoSteer-$6$K-PT & EgoSteer-$9.6$K-PT & EgoSteer-$0$K-FT & EgoSteer-$3$K-FT & EgoSteer-$6$K-FT & EgoSteer-$9.6$K-FT \\
\midrule
Camera setup            & Head & Head & Head & Head \& Chest & Head \& Chest & Head \& Chest & Head \& Chest \\
Resolution              & $384\times384$ & $384\times384$ & $384\times384$ & $640\times480$ & $640\times480$ & $640\times480$ & $640\times480$ \\
GPUs                    & 64 A800 & 64 A800 & 128 A800 & 64 A800 & 32 A800 & 64 A800 & 64 A800 \\
Gradient accumulation   & 2 & 4 & 2 & 2 & 4 & 2 & 2 \\
Global batch size       & 2304 & 4608 & 4608 & 512 & 512 & 512 & 512 \\
Training steps          & 100K & 100K & 160K & 60K & 60K & 60K & 60K \\
Learning rate (VLM / AE / WM) & \multicolumn{3}{c}{$1{\times}10^{-4}$ / $3{\times}10^{-4}$ / $3{\times}10^{-4}$}
                          & \multicolumn{4}{c}{$1{\times}10^{-5}$ / $3{\times}10^{-5}$ / $3{\times}10^{-5}$} \\
\bottomrule
\end{tabular}}
\vspace{8pt}

\end{table}

\begin{table}[!htbp]
\caption{Per-task success rates for the pre-training scaling study in \Cref{subsec:scaling-and-baseline}. Each task is evaluated over 10 randomized trials. Bold values highlight the best performance.}
\label{tab:per-task-sr}
\centering
\small
\setlength{\tabcolsep}{6pt}
\begin{tabular}{lcccc}
\toprule
Task & EgoSteer-$0$K-FT & EgoSteer-$3$K-FT & EgoSteer-$6$K-FT & EgoSteer-$9.6$K-FT \\
\midrule
Grasp object               & 80\% & 80\% & 70\% & \textbf{100\%} \\
Hand over object           & 70\% & 80\% & 80\% & \textbf{100\%} \\
Place items into container & 40\% & 80\% & 90\% & \textbf{100\%} \\
Point at object            & 20\% & 40\% & 60\% & \textbf{70\%}  \\
Place toy chick into slot  & 30\% & 0\%  & 20\% & \textbf{40\%}  \\
Pull out tissue            & 60\% & 20\% & 50\% & \textbf{80\%}  \\
Push ball into box         & 0\%  & 0\%  & 10\% & \textbf{20\%}  \\
Put garbage into trash bin & 0\%  & \textbf{60\%} & 30\% & 30\%  \\
Stack paper cups           & 0\%  & 20\% & 10\% & \textbf{40\%}  \\
Stack tableware            & 0\%  & \textbf{20\%} & 10\% & \textbf{20\%}  \\
\midrule
\textbf{Average} & 30\% & 40\% & 43\% & \textbf{60\%} \\
\bottomrule
\end{tabular}
\vspace{8pt}

\end{table}

\begin{table}[t]
\caption{Post-training configurations of $\pi_{0.5}$ and Being-H0.5.
All methods are post-trained on the same real-robot dataset for
60K optimizer updates and evaluated using their final checkpoints.
Both baselines retain their officially recommended input resolution.
For $\pi_{0.5}$, we adopt EgoSteer's 48-dimensional wrist-pose and
fingertip representation.
For Being-H0.5, we retain the officially recommended action representation.}
\label{tab:baseline-post-training}
\centering
\small
\setlength{\tabcolsep}{6pt}
\renewcommand{\arraystretch}{1.08}
\begin{tabular}{lcc}
\toprule
Hyperparameter
    & $\pi_{0.5}$
    & Being-H0.5 \\
\midrule
Pre-training checkpoint
    & \texttt{pi05\_droid}
    & Official released base checkpoint \\
Camera setup
    & Head \& Chest
    & Head \& Chest \\
Resolution
    & $224\times224$
    & $224\times224$ \\
Action representation
    & \begin{tabular}[c]{@{}c@{}}
        Wrist poses + \\
        fingertip positions
      \end{tabular}
    & \begin{tabular}[c]{@{}c@{}}
        EEF poses + \\
        dexterous hand joints
      \end{tabular} \\
Gradient accumulation
    & 1
    & 1 \\
Global batch size
    & 128
    & 672 \\
Training steps
    & 60K
    & 60K \\
Peak learning rate
    & $5\times10^{-5}$
    & $1\times10^{-4}$ \\
Freeze-VLM steps
    & 0
    & 0 \\
Warmup steps
    & 1,500
    & 3,000 \\
Evaluation checkpoint
    & Final at 60K steps
    & Final at 60K steps \\
\bottomrule
\end{tabular}
\end{table}

\begin{table}[!htbp]
\caption{Per-task comparison between EgoSteer-$9.6$K-FT and two VLA baselines, Being-H0.5 and $\pi_{0.5}$, in \Cref{subsec:scaling-and-baseline}. All methods are fine-tuned on our real-robot dataset and evaluated on the same 10 tasks. Each task uses 10 randomized trials. Bold values highlight the best performance.}
\label{tab:cmp-baselines}
\centering
\small
\setlength{\tabcolsep}{6pt}
\begin{tabular}{lccc}
\toprule
Task & EgoSteer-$9.6$K-FT & Being-H0.5 & $\pi_{0.5}$ \\
\midrule
Grasp object                       & \textbf{100\%} & 80\%          & 80\%          \\
Hand over object                   & \textbf{100\%} & 60\%          & 20\%          \\
Place items into container         & \textbf{100\%} & 50\%          & 0\%           \\
Pour items out of box              & \textbf{50\%}  & 30\%          & 0\%           \\
Place bread on tray                & 70\%           & \textbf{80\%} & 40\%          \\
Pull out tissue                    & \textbf{80\%}  & 10\%          & 30\%          \\
Place item at specific orientation & \textbf{30\%}  & \textbf{30\%} & \textbf{30\%} \\
Attach eraser to whiteboard        & \textbf{90\%}  & 50\%          & 20\%          \\
Put garbage into trash bin         & \textbf{30\%}  & 0\%           & 0\%           \\
Put tennis ball into bucket        & \textbf{90\%}  & 0\%           & 0\%           \\
\midrule
\textbf{Average} & \textbf{74\%} & 39\% & 22\% \\
\bottomrule
\end{tabular}
\vspace{8pt}

\end{table}

\begin{table}[!htbp]
\caption{Per-task success rates for the ablation study in \Cref{subsec:ablation}. EgoSteer-$1$K-FT is compared with  ablated variants. Each task uses 10 randomized trials. Bold values highlight the best performance.}
\label{tab:cmp-ablation}
\centering
\small
\setlength{\tabcolsep}{6pt}
\resizebox{0.85\textwidth}{!}{%
\begin{tabular}{lcccc}
\toprule
Task & EgoSteer-$1$K-FT & No WM-objective & No training-RTC & Noisy data \\
\midrule
Grasp object                       & \textbf{60\%} & 40\%          & 30\%          & \textbf{60\%} \\
Hand over object                   & \textbf{40\%} & 20\%          & 30\%          & \textbf{40\%} \\
Place items into container         & 50\%          & 30\%          & 50\%          & \textbf{70\%} \\
Pour items out of box              & \textbf{60\%} & 50\%          & 10\%          & 0\%           \\
Place bread on tray                & \textbf{70\%} & 10\%          & 60\%          & 50\%          \\
Pull out tissue                    & 20\%          & 30\%          & \textbf{60\%} & 10\%          \\
Place item at specific orientation & \textbf{20\%} & 10\%          & 10\%          & \textbf{20\%} \\
Attach eraser to whiteboard        & 50\%          & \textbf{80\%} & 50\%          & 40\%          \\
Put garbage into trash bin         & \textbf{30\%} & 0\%           & 0\%           & 0\%           \\
Put tennis ball into bucket        & 40\%          & 40\%          & \textbf{90\%} & 40\%          \\
\midrule
\textbf{Average} & \textbf{44\%} & 31\% & 39\% & 33\% \\
\bottomrule
\end{tabular}}
\vspace{8pt}

\end{table}

\begin{table}[!htbp]
\caption{Training hyperparameters for the ablation study in \Cref{subsec:ablation}. All variants are pre-trained on 1K hours of egocentric data and fine-tuned on the same real-robot dataset, without DAgger refinement. Pre-training steps are selected at the lowest evaluation $L_1$ loss for each pre-training run.}
\label{tab:ablation-training-config}
\centering
\small
\setlength{\tabcolsep}{5pt}
\renewcommand{\arraystretch}{1.15}
\resizebox{\linewidth}{!}{%
\begin{tabular}{l cccc @{\hskip 14pt} cccc}
\toprule
 & \multicolumn{4}{c}{Pre-Training} & \multicolumn{4}{c}{Fine-Tuning} \\
\cmidrule(lr){2-5} \cmidrule(lr){6-9}
Hyperparameter & EgoSteer-$1$K-PT & No WM-objective & No training-RTC & Noisy data & EgoSteer-$1$K-FT & No WM-objective & No training-RTC & Noisy data \\
\midrule
Camera setup            & Head & Head & Head & Head & Head & Head & Head & Head \\
Resolution              & $384\times384$ & $384\times384$ & $384\times384$ & $384\times384$ & $384\times384$ & $384\times384$ & $384\times384$ & $384\times384$ \\
GPUs                    & 64 A800 & 64 A800 & 64 A800 & 64 A800 & 64 A800 & 64 A800 & 64 A800 & 64 A800 \\
Global batch size       & 1152 & 1152 & 1152 & 1152 & 1152 & 1152 & 1152 & 1152 \\
Training steps          & 30K & 80K & 30K & 20K & 60K & 60K & 60K & 60K \\
Learning rate (VLM / AE / WM) & \multicolumn{4}{c}{$1{\times}10^{-4}$ / $3{\times}10^{-4}$ / $3{\times}10^{-4}$}
                          & \multicolumn{4}{c}{$1{\times}10^{-5}$ / $3{\times}10^{-5}$ / $3{\times}10^{-5}$} \\
\bottomrule
\end{tabular}}
\vspace{8pt}

\end{table}

\begin{table}[t]
\caption{Few-shot fine-tuning hyperparameters of EgoSteer-$9.6$K-FS for the five long-horizon dexterous tasks in \Cref{subsec:single-task}.}
\label{tab:dexterous-single-task}
\centering
\small
\setlength{\tabcolsep}{6pt}
\renewcommand{\arraystretch}{1.08}
\begin{tabular}{@{}ccc@{}}
\toprule
Hyperparameter
    & \multicolumn{2}{c}{Shared value} \\
\midrule
Pre-training checkpoint
    & \multicolumn{2}{c}{EgoSteer-$9.6$K-PT at 170K steps} \\
Camera setup
    & \multicolumn{2}{c}{Head} \\
Resolution
    & \multicolumn{2}{c}{$384\times384$} \\
GPUs
    & \multicolumn{2}{c}{8 A800} \\
Gradient accumulation
    & \multicolumn{2}{c}{1} \\
Global batch size
    & \multicolumn{2}{c}{144} \\
Learning rate (VLM / AE / WM)
    & \multicolumn{2}{c}{$1{\times}10^{-5}$ / $3{\times}10^{-5}$ / $3{\times}10^{-5}$} \\
\midrule
Task
    & \multicolumn{2}{c}{Fine-tuning steps} \\
\cmidrule(lr){2-3}
    & 20 demonstrations & 100 demonstrations \\
\midrule
Box-Folding   & $1.6 \times 10 ^ 4$ & $2.4 \times 10 ^ 4$ \\
Cake-Unboxing & $3.5 \times 10 ^ 4$ & $2.3 \times 10 ^ 4$ \\
Cup-Stacking  & $1.3 \times 10 ^ 4$ & $1.9 \times 10 ^ 4$ \\
Phone-Packing & $1.6 \times 10 ^ 4$ & $2.0 \times 10 ^ 4$ \\
Cola-Making   & $1.8 \times 10 ^ 4$ & $2.6 \times 10 ^ 4$ \\
\bottomrule
\end{tabular}
\end{table}

\clearpage

\end{document}